\documentclass{ecai} 
\usepackage{caption} 

\usepackage{enumitem}

\usepackage{subfig}

\usepackage{algorithm}
\usepackage{algpseudocode}

\usepackage{amsmath}
\usepackage{mathtools}
\usepackage{amsthm}
\usepackage{xparse}
\usepackage{xspace}
\usepackage{latexsym}
\usepackage{amssymb}
\usepackage{booktabs}
\usepackage{enumitem}
\usepackage{graphicx}
\usepackage{color}
\usepackage{pifont}
\usepackage{tabularx}
\usepackage{threeparttable}
\usepackage{makecell}
\usepackage{multicol}
\usepackage{multirow}
\usepackage{arydshln}
\usepackage{appendix}
\usepackage{graphicx} 

\newcommand{\BibTeX}{B\kern-.05em{\sc i\kern-.025em b}\kern-.08em\TeX}

\begin{document}


\begin{frontmatter}


\paperid{123} 


\title{FedSODA: Federated Fine-tuning of LLMs via Similarity Group Pruning and Orchestrated Distillation Alignment}


\author[A]{\fnms{Manning}~\snm{Zhu}}
\author[A]{\fnms{Songtao}~\snm{Guo}\thanks{Corresponding Author. Email: guosongtao@cqu.edu.cn}}
\author[A]{\fnms{Pengzhan}~\snm{Zhou}} 
\author[B]{\fnms{Yansong}~\snm{Ning}} 
\author[A]{\fnms{Chang}~\snm{Han}}
\author[C]{\fnms{Dewen}~\snm{Qiao}}
\address[A]{Chongqing University}
\address[B]{The Hong Kong University of Science and Technology (Guangzhou)	}
\address[C]{Third Military Medical University}


\begin{abstract}
Federated fine-tuning (FFT) of large language models (LLMs) has recently emerged as a promising solution to enable domain-specific adaptation while preserving data privacy.
Despite its benefits, FFT on resource-constrained clients relies on the high computational and memory demands of full-model fine-tuning, which limits the potential advancement.
This paper presents FedSODA, a resource-efficient FFT framework that enables clients to adapt LLMs without accessing or storing the full model.
Specifically, we first propose a similarity group pruning (SGP) module, which prunes redundant layers from the full LLM while retaining the most critical layers to preserve the model performance. 
Moreover, we introduce an orchestrated distillation alignment (ODA) module to reduce gradient divergence between the sub-LLM and the full LLM during FFT. 
Through the use of the QLoRA, clients only need to deploy quantized sub-LLMs and fine-tune lightweight adapters, significantly reducing local resource requirements. 
We conduct extensive experiments on three open-source LLMs across a variety of downstream tasks. 
The experimental results demonstrate that FedSODA reduces communication overhead by an average of \textbf{70.6\%}, decreases storage usage by \textbf{75.6\%}, and improves task accuracy by \textbf{3.1\%}, making it highly suitable for practical FFT applications under resource constraints. 

\end{abstract}

\end{frontmatter}


\section{Introduction}
Large language models (LLMs) have demonstrated remarkable performance on various tasks, such as chatbots~\cite{chatrobot}, legal consulting \cite{law1,law2} and medical diagnostics \cite{medical1,medical2,medical3}. However, adapting LLMs to specific downstream tasks typically requires fine-tuning the entire model, which demands substantial computational resources and sufficient data tailored to the target task. In many real-world applications, data is distributed across different institutions, and due to privacy and regulatory constraints, data sharing among these entities is often strictly limited. To efficiently fine-tune LLMs while preserving data privacy, federated learning (FL) \cite{mcmahan:2016federated,mcmahan:2017communication} has emerged as a critical approach. FL enables distributed clients to collaboratively fine-tune LLMs while ensuring that data remains local and private. In this work, we focus on federated fine-tuning (FFT) of LLMs, aiming to allow distributed clients to collaboratively adapt LLMs to downstream tasks under privacy-preserving conditions.

\begin{figure}
    \centering
    \includegraphics[width=\linewidth]{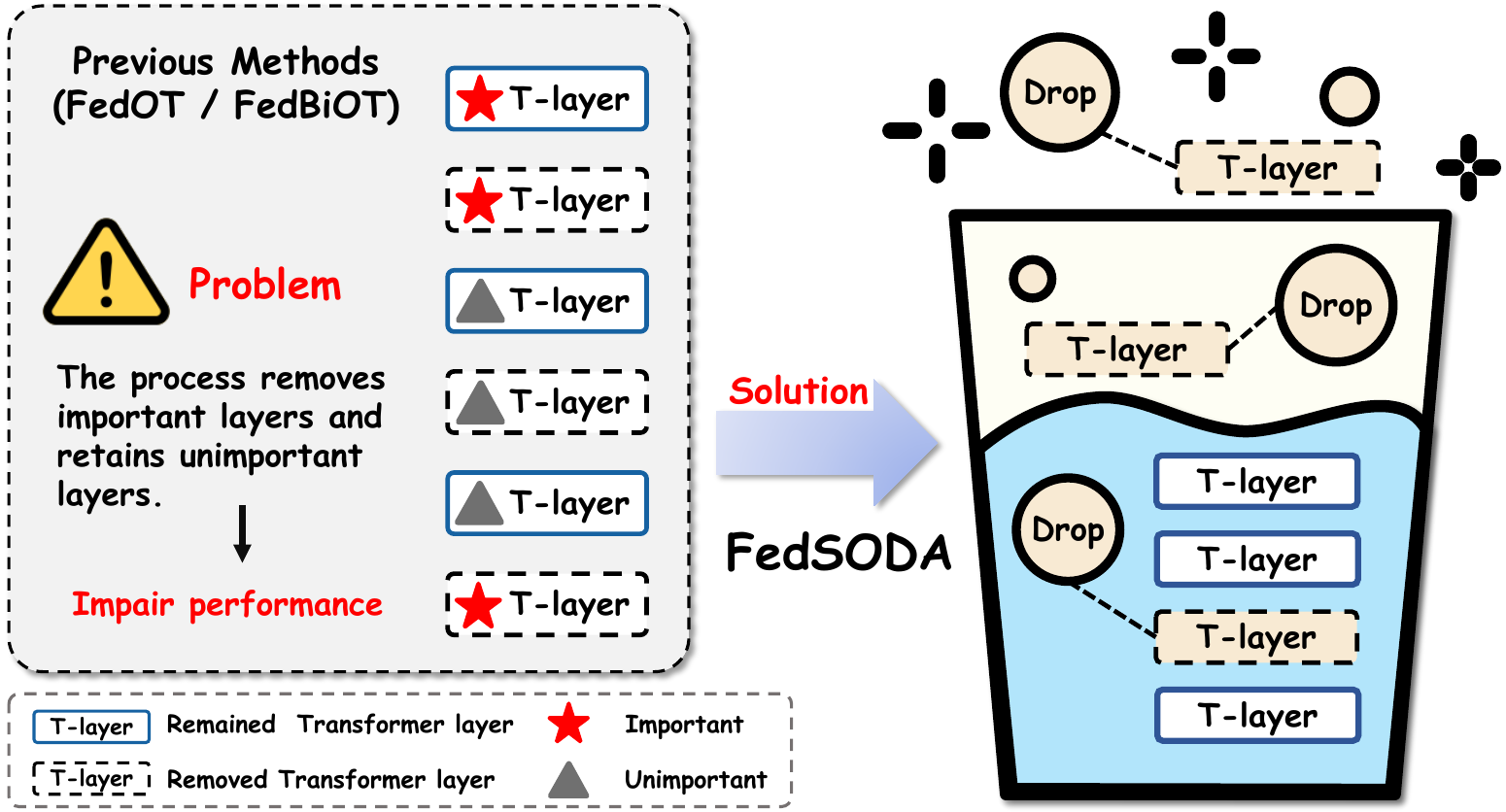}
    \vspace{0.5px}
    \caption{FedSODA improves performance by selectively retaining important layers and removing unimportant layers.}
    \vspace{10px}
    \label{fig1}
\end{figure}

While FFT of LLMs offers significant advantages, directly fine-tuning all model parameters is computationally expensive and incurs high communication costs, making it impractical. To address this, existing studies \cite{fedpft,fedprompt,FedPepTAO,fedpetuning} have introduced parameter-efficient fine-tuning (PEFT)~\cite{peft} methods into the FFT. PEFT reduces communication and computation overhead by freezing most of the model parameters and only fine-tuning a small portion or newly introduced parameters. However, even with these methods, the gradient computation of the tunable parameters still relies on backpropagation through the entire LLM~\cite{lst}, which limits further efficiency improvements.

Therefore, recent approaches, such as Offsite-tuning~\cite{offsite-tuning} and FedBiOT~\cite{fedbiot}, attempt to fine-tune LLMs without requiring access to the full model. These methods leverage techniques like layer dropout~\cite{layerdrop} to compress the full LLM, creating smaller and more fine-tuning-friendly sub-LLMs. This eliminates the need for backpropagation through the entire LLM. 
Nevertheless, the aggressive pruning of transformer layers in Offsite-tuning and FedBiOT discards critical model information, as illustrated on the left of Figure~\ref{fig1}, which may hinder the sub-LLMs' abilities to approximate the outputs of the full LLM. Offsite-tuning further leads to gradient error accumulation in the sub-LLM relative to the full LLM, resulting in a sharp drop in performance. 
In addition, FedBiOT’s iterative alignment of sub-LLMs with the full LLM introduces significant overhead in both communication and training time, ultimately limiting the improvements in FFT training speed and resource utilization.

To address the aforementioned limitation, we propose \textbf{Fed}erated Fine-tuning of LLMs via \textbf{S}imilarity Group Pruning and \textbf{O}rchestrated \textbf{D}istillation \textbf{A}lignment (\textbf{FedSODA}), a lightweight yet powerful framework for efficient FFT of LLMs as shown on the right of Figure~\ref{fig1}.
Specifically, we first follow the Offsite-tuning architecture to decompose the transformer layers of the full LLM into two independent modules: the emulator and the adapter.
Then, we introduce similarity group pruning (SGP), which compresses the original emulator into a smaller sub-emulator through identifying and pruning unimportant layers based on representational similarity. 
By strategically removing layers with similarity, SGP achieves significant model compression while preserving the essential knowledge embedded within the emulator.
Moreover, we propose orchestrated distillation alignment (ODA), which performs intermittent, partial-layer alignment between the full LLM and the compressed LLM (sub-LLM) before and during the FFT process. 
Through this periodic alignment mechanism, ODA effectively minimizes gradient discrepancies between the full LLM and the sub-LLM, consequently reducing the required number of alignment rounds and parameter transmission requirements during FFT. 
Finally, FedSODA employs QLoRA~\cite{qlora} during the fine-tuning phase on client devices to further enhance communication efficiency by drastically reducing transmission requirements between clients and the server.

To validate the effectiveness of our proposed framework, we conducted experiments using diverse base models including LLaMA3-8B~\cite{touvron2024llama3}, Qwen2-7B~\cite{baichuan2024qwen2}, and Mistral-7B~\cite{mistral2023mistral7b}. The empirical results demonstrate significant advantages for FedSODA: it reduces memory consumption by an average of \textbf{75.6\%} compared to baseline methods, a reduction substantial enough to enable the fine-tuning of a 7B-parameter model on a single consumer-grade GPU. 
Furthermore, FedSODA achieves a substantial reduction in communication overhead, with an average of \textbf{70.6\%} fewer parameters transmitted during FFT. 
Beyond efficiency gains, our framework also improves downstream task performance by an average of \textbf{3.1\%} across multiple datasets, encompassing commonsense and mathematical reasoning benchmarks, uncovering its effectiveness and practical applicability, especially in resource-constrained environments.

Our main contributions are mainly four-fold:
\vspace{-5px}
\begin{itemize}[leftmargin=2em]
\item We propose FedSODA, an efficient federated fine-tuning (FFT) framework for LLMs, designed to address key limitations in FFT systems, specifically the constraints of client-side resources and the high communication overhead inherent in FFT.

\item We design the similarity group pruning (SGP) module, which selectively prunes redundant layers in the emulator based on representational similarity. This retains core knowledge while significantly reducing model size, overcoming the information loss issues found in prior layer dropout methods.

\item We introduce orchestrated distillation alignment (ODA) module, a new alignment mechanism that performs intermittent partial layer alignment during training. This greatly reduces communication overhead and fine-tuning time, while mitigating gradient error accumulation in the sub-LLM relative to the full LLM.

\item We conduct extensive experiments and the results demonstrate that our method significantly improves memory efficiency, communication overhead, and task performance compared to baseline approaches.  
\end{itemize}

\section{Related Works}\label{sec:related_work}
\subsection{Federated Learning}

FL is a decentralized machine learning paradigm that allows multiple clients to collaboratively train models while preserving client data privacy \cite{mcmahan:2016federated,mcmahan:2017communication}. However, the non-independent and identically distributed (Non-IID) nature of client data often leads to performance degradation. Recent studies have shown that fine-tuning LLMs on Non-IID data is less sensitive to performance loss, providing valuable insights into how LLMs can be effectively leveraged within the FL framework \cite{chen2023,nguyen2022}. However, fine-tuning LLM in FL suffers from communication and computational bottlenecks, which limits its practical applications.

\subsection{Federated Fine-Tuning of LLMs with PEFT}
In federated fine-tuning (FFT) of LLMs, parameter-efficient fine-tuning (PEFT) methods \cite{peft} address the challenges of high computational and communication costs by freezing most of the pre-trained model parameters and only fine-tuning a small subset or newly introduced parameters. By minimizing the number of parameters to be updated and communicated, PEFT enables more efficient model updates in FFT. Examples of such methods include FedPrompt \cite{fedprompt}, which aggregates prompts to minimize communication, and FedPepTAO \cite{FedPepTAO}, which employs adaptive optimization to select appropriate prompt layers for FFT. Despite these efforts, the gradient computation for tunable parameters still needs backpropagation through the full LLM \cite{lst}. Recently, Offsite-tuning \cite{offsite-tuning} utilizes the compressed emulator to approximate the behavior of the full LLM, providing gradients for fine-tuning the adapter, and reducing reliance on the full LLM. FedOT \cite{federatedscope} extends this concept to FL. Additionally, FedBiOT \cite{fedbiot} introduces bi-level optimization, alternating updates between the emulator and adapter, to further enhance privacy while improving performance. All three methods use public datasets to approximate the full LLM through the emulator, thus avoiding fine-tuning the full LLM. 

However, these methods either ignore the intrinsic characteristics of a large number of intermediate layers in LLMs or face the problem of gradient error accumulation~\cite{fedpft}, resulting in poor fine-tuning performance. In addition, these methods still face huge memory challenges, making it difficult for resource-constrained clients to load compressed models. Unlike traditional methods, we construct sub-LLM based on similarity pruning and mitigate gradient error accumulation through a two-phase distillation. Meanwhile, we use quantization methods to enable sub-LLMs to be lightweight and deployed on resource-constrained clients.

\section{Preliminary} \label{sec:preliminary}

\subsection{LLM Fine-tuning with QLoRA}

\textbf{Quantized LoRA (QLoRA)} \cite{qlora} integrates both Low-Rank Adaptation (LoRA) \cite{lora} and 4-bit NormalFloat Quantization (NF4). During fine-tuning, the LLM parameters \(\Theta = \{w_1, w_2, \dots, w_N\}\) remain frozen, and only two low-rank matrices \(A\) and \(B\) are trained. Moreover, NF4 quantization is applied to both \(\Theta\) and the matrices \(A, B\), substantially reducing memory usage while preserving the statistical properties of the high-precision model.

For a high-precision weight \(X^{\mathrm{HP}}\), the quantization process is:
\begin{equation}
X^{\mathrm{INT}} = \mathrm{round}\left[(2^N - 1) \cdot \Phi\bigl(\tfrac{X^{\mathrm{HP}}}{\sigma}\bigr)\right],
\end{equation}
where $N$ refers to $N$-bit quantization, \(\Phi(\cdot)\) is the standard normal cumulative distribution function (CDF), and \(\sigma\) denotes the standard deviation computed from the weights.
\begin{figure*}
    \centering
    \includegraphics[width=\linewidth]{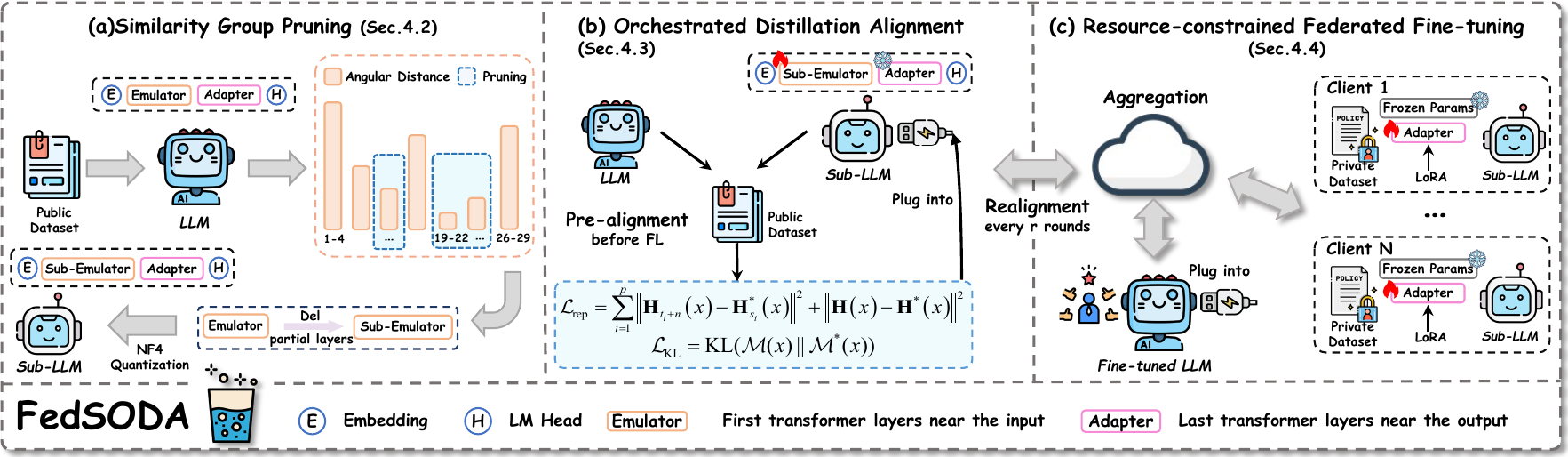}
    \vspace{-5px}
    \caption{The overall framework of FedSODA that consists of three key modules that enhance the adaptation of full LLM to downstream tasks through FFT: (a) constructing a sub-LLM using the similarity group pruning module; (b) enabling the sub-LLM to mimic the performance of the full LLM and reduce gradient errors through the orchestrated distillation alignment module; (c) performing local fine-tuning of the lightweight adapter in the sub-LLM using QLoRA.}
    \label{fig2}
    \vspace{5px}
\end{figure*}

During fine-tuning, the model parameters are updated as:
\begin{equation}
\begin{gathered}
\Theta^{\prime} = \Theta + \Delta \Theta; \\
\Delta \Theta = \mathop{\arg\min}\limits_{\Delta \Theta} L(\Theta + \Delta \Theta, D),
\end{gathered}
\end{equation}
where \(D\) denotes the downstream dataset, and only \(\Delta \Theta=BA\) is optimized, while \(\Theta\) itself remains frozen.  \(A \in \mathbb{R}^{r \times k}\) and \(B \in \mathbb{R}^{d \times r}\) with \(r \ll \min(d, k)\), are initialized as follows:
\begin{equation}
   A \sim \mathcal{N}(0, \sigma_A^2), \quad B = 0, 
\end{equation}
where $\mathcal{N}(0, \sigma_A^2)$ denotes a normal distribution with mean 0 and variance $\sigma_A^2$, and $B$ is initialized to zero.

\subsection{Problem Definition}

In our framework, the central server is equipped with ample computational and storage resources to host a pre-trained \textbf{full LLM} \( \mathcal{M} \). We follow the Offsite-tuning \cite{offsite-tuning} setting and divide transformer layers of the full LLM into two key components: an \textbf{emulator} (\( \mathcal{E} \)) and an \textbf{adapter} (\( \mathcal{A} \)), such that \( \mathcal{M} = [\mathcal{E}, \mathcal{A}] \). The adapter consists of the last \( L_{\mathcal{A}} \) layers of the LLM, while the emulator contains the remaining \( L_{\mathcal{E}} \) layers. The full LLM parameters are denoted by \( \Theta = \{w_{\mathcal{E}}, w_{\mathcal{A}}\} \), where \( w_{\mathcal{E}} \) and \( w_{\mathcal{A}} \) are the parameters of the emulator and the adapter, respectively.

Next, we apply a pruning operation \( C(\cdot) \) to the emulator \( \mathcal{E} \) to reduce the model size, resulting in a \textbf{sub-emulator} \( \mathcal{E}^* = C(\mathcal{E}) \). The sub-emulator is then combined with the adapter to form a \textbf{sub-LLM} \( \mathcal{M}^* = [\mathcal{E}^*, \mathcal{A}] \), which is deployed on the central server and resource-constrained clients \( C_k \), where \( 1 \leq k \leq N \). The parameters of the sub-LLM are denoted as \( \Theta^* =  \{w_{\mathcal{E}^*}, w_{\mathcal{A}}\} \). 

Subsequently, the sub-LLM's adapter \( \mathcal{A} \) is fine-tuned on each client \( C_k \) using its local dataset \(D_k\), minimizing the global objective function \( L ( \Theta^* + \Delta \Theta_{\mathcal{A}}^* ) \). After fine-tuning, an approximate value \( \Delta \Theta_{\mathcal{A}}^* \) of \( \Delta \Theta_{\mathcal{A}} \) is obtained and plugged back into the full LLM, resulting in the updated parameters \( \Theta^{'} \). The process can be formally described as:

\begin{equation}
\begin{gathered}
\Theta^{\prime} = \Theta + \Delta \Theta_{\mathcal{A}}^*; \\
\Delta \Theta_{\mathcal{A}}^* = \mathop{\arg\min}\limits_{\Delta \Theta_{\mathcal{A}}^*}
\sum_{k=1}^N \frac{\lvert D_{k} \rvert}{\sum_{k=1}^N \lvert D_k \rvert}
L(\Theta^* + \Delta \Theta_{\mathcal{A}}^*, D_k),
\end{gathered}
\end{equation}
\section{Methodology} \label{sec:methodology}
\subsection{Framework Overview}

The overall framework of our proposed approach, named FedSODA, is depicted in Figure~\ref{fig2}. For a detailed implementation, the complete pseudocode of FedSODA can be found in Appendix \ref{app:alo}. The core objective of our framework is to enable resource-constrained local clients to fine-tune the full LLM by fine-tuning the sub-LLM, achieving gradients that approximate those of the full LLM, thereby allowing the full LLM to effectively adapt to downstream tasks.
\textbf{First}, we activate the similarity group pruning (SGP) module, which utilizes the public dataset \( D_{\text{public}} \) to perform SGP on the emulator \( \mathcal{E} \) within the full LLM $(\mathcal{M} = [\mathcal{E}, \mathcal{A}])$. 
The use of the public dataset is motivated by practical constraints and is consistent with common practice in prior work \cite{fedpft,fedbiot,federatedscope}. We provide a detailed discussion of this design choice in Appendix~\ref{app:public datasets}.
This pruning operation generates a sub-emulator \( \mathcal{E}^* \) with reduced complexity while retaining core functionality (referred to Section~\ref{compression}). 
 The sub-emulator is combined with the adapter to form the sub-LLM \( (\mathcal{M}^* = [\mathcal{E}^*, \mathcal{A}]) \). However, the sub-LLM remains challenging to deploy on resource-constrained clients. To address this, we apply NF4 quantization to the sub-LLM and add LoRA modules to selected layers of the sub-emulator and all layers of the adapter.
\textbf{Next}, we employ the orchestrated distillation alignment (ODA) module, where knowledge distillation is performed in two phases: prior to federated fine-tuning (FFT) and during certain rounds of FFT on the server. This provides a solid foundation for FFT and minimizes the gradient errors of the sub-LLM (see Section~\ref{alignment}). \textbf{Subsequently}, each client \( C_k \) receives the NF4-quantized sub-LLM. \( \mathcal{A}_k \) is fine-tuned on the client's private dataset \( D_k \), while the sub-emulator remains frozen to minimize computational overhead. After local training, clients upload their adapter's LoRA weights to the server, which aggregates these weights to update the global model parameters. The aggregated parameters are then distributed back to the clients (see Section~\ref{FL Fine-Tuning}). 
\textbf{Finally}, the fine-tuned adapter LoRA modules from the sub-LLM \textbf{plug into} the full LLM, enabling the full LLM to adapt effectively to downstream tasks.
\vspace{-10px}
\subsection{Similarity Group Pruning}
\label{compression}
\begin{algorithm}
    \caption{Similarity Group Pruning}
    \label{alg.Layer prune}
    \begin{flushleft}
    \textbf{Input}: Emulator $\mathcal{E}$ with $L_\mathcal{E}$ layers, adapter $\mathcal{A}$, dataset $D$, pruning configuration specifying $n$ pruned layers per group and $p$ groups in total.\\
    \textbf{Output}: Sub-LLM $\mathcal{M}^*$.
    \end{flushleft}
    \begin{algorithmic}[1]
    \State \textbf{initialize} distances $\leftarrow$ [ ]
    \For{$\ell = 0$ \textbf{to} $(L_\mathcal{E} - n)$}
        \State $\text{distance} \leftarrow$ \textbf{COSINE\_DISTANCE}($x_{\ell}$, $x_{\ell + n}$) \Comment{calculate the angular distance between layer $\ell$ and layer $\ell+n$ on dataset $D$}
        \State \textbf{append} $(\ell,\ell+n,\text{distance} )$ to distances
    \EndFor
    \State pruned\_groups $\leftarrow$ \textbf{Find\_Min\_Distance}(distances, p) 
    \Comment{find the optimal non-overlapping $p$ groups with the minimal sum of angular distances}
    \State $\mathcal{E^*} \leftarrow$ Prune\_Layers($\mathcal{E}$, pruned\_groups) 
    \Comment{preserve the first layer and remove the next $n$ layers in each pruned group}
    \State   $\mathcal{M}^*\leftarrow [\mathcal{E}^*, \mathcal{A}]$
    \State \textbf{return} 
    $\mathcal{M}^*$
    \end{algorithmic}
          
\end{algorithm}

In transformer models, residual connections allow the final output to be expressed as a sum of all previous layer outputs and the input embedding~\cite{attention}. This additive structure makes it possible to prune layers that contribute least to the final output—specifically, those showing minimal change across adjacent layers. However, prior methods such as Offsite-tuning and FedBiOT may mistakenly remove important layers, leading to performance drops, as illustrated on the left of Figure~\ref{fig1}.

To address this, we propose an efficient similarity group pruning (SGP) module for emulator compression, inspired by \cite{unreasonable}. SGP groups transformer layers and evaluates their similarity using angular distance, a metric well-suited for high-dimensional data. Unlike Euclidean distance and KL divergence---which suffer from sensitivity to magnitude, instability with zeros, and high computational cost---angular distance emphasizes direction, aligning naturally with the directional nature of residual connections. By pruning groups with the highest similarity, SGP reduces model complexity while preserving performance. The full algorithm is presented in Algorithm~\ref{alg.Layer prune}.

We utilize a neutral dataset or a dataset pertinent to downstream tasks \(D_{\text{public}}\) for this process. Specifically, we group every \( n+1 \) layers and compute the similarity between layer \( \ell \) and layer \( \ell+n \), where \( n \) denotes the number of layers to prune in each group. The similarity between layer \( \ell \) and layer \( \ell+n \) is measured using the angular distance, defined as:
\begin{equation}
    d(x^{(\ell)}, x^{(\ell+n)}) \equiv \frac{1}{\pi} \arccos{\frac{x_S^{(\ell)} \cdot x_S^{(\ell+n)}}{\|x_S^{(\ell)}\| \|x_S^{(\ell+n)}\|}},
    \label{angular_distance}
\end{equation}
where \( x_S^{(\ell)} \) represents the hidden state of the model at layer \( \ell \) with an input sequence length of $S$, ``\( \cdot \)'' denotes the dot product, and \( \| \cdot \| \) denotes the \( l_2 \)-norm. The angular distance is then aggregated across multiple samples to obtain a robust similarity measure.

After calculating the angular distances for all groups, we select the top \( p \) non-overlapping groups with the smallest summed angular distances. This selection ensures that the pruning operation minimally impacts the overall model performance. Consequently, we prune layers from the second to the \( n+1 \)-th layer in each selected group, resulting in a pruned sub-emulator \( \mathcal{E}^* \) with \( L_{\mathcal{E}^*}\) layers, where \(L_{\mathcal{E}^*} =  L_{\mathcal{E}} - n \times p \). 

Additionally, we preserve the starting layer indices of each pruned group and maintain an index mapping \(  \mathcal{I} \) between the starting layers in the original emulator and the sub-emulator.

Overall, SGP provides a principled and efficient mechanism for compressing transformer models, facilitating their deployment in resource-constrained environments without compromising performance.

\vspace{-10px}
\subsection{Orchestrated Distillation Alignment}
\label{alignment}

The SGP module (Sec.~\ref{compression}) compresses the original emulator \( \mathcal{E} \) into a sub-emulator \( \mathcal{E}^* \) by removing \( p \) groups of \( n \) layers. 
Compression causes output discrepancies in \( \mathcal{M}^* \), leading to inaccurate gradients for the full LLM \( \mathcal{M} = [\mathcal{E}, \mathcal{A}] \). To address this, we propose the orchestrated distillation alignment (ODA) module, which reduces output deviation through a two-phase knowledge distillation process, improving gradient quality for FFT.

\textbf{ODA consists of two phases:}

\textit{Pre-alignment:} Before FFT, the sub-LLM aligns with the full LLM by knowledge distillation from retained layers in the pruned \( p \) groups. This ensures that \( \mathcal{M}^* \) approximates \( \mathcal{M} \)'s representations, reducing gradient error from pruning.

\textit{Realignment:} During FFT, the sub-LLM may drift due to adapter updates. To counteract this, periodic alignment (every \( r \) rounds) is performed using randomly sampled data from \( D_{\text{public}} \), balancing alignment and task adaptation.

We define layer mappings as \( \mathcal{I}_i = (t_i, s_i), i \in [1, p] \), where \( t_i \) is the start of the \( i \)-th pruned group in \( \mathcal{M} \), and \( s_i \) is the corresponding layer in \( \mathcal{M}^* \). During alignment, adapter parameters \( w_{\mathcal{A}} \) and sub-emulator weights \( w_{\mathcal{E}^*} \) are frozen. Only LoRA modules at \( s_i \) are trainable, with parameters \( w_{\mathcal{E}^*_{\text{lora}}} \).

We ensure that the output of layer \( s_i \) in the sub-LLM closely aligns with the output at \( t_i \) in the full LLM, while the final outputs of both models remain approximately equivalent. To this end, we aim to minimize the following $l_2$-norm:

\begin{equation}
\begin{gathered}
\mathcal{L}_{\text{rep}} = \mathcal{L}_{\text{inter}} + \mathcal{L}_{\text{final}}; \\
\mathcal{L}_{\text{inter}} = \sum_{i=1}^{p} \left\| \textbf{H}_{t_i+n}\left(x; w_{\mathcal{E}}\right) - \textbf{H}_{s_i}^*\left(x; w_{\mathcal{E}^{*\prime}}\right) \right\|^2; \\
\mathcal{L}_{\text{final}} = \left\| \textbf{H}\left(x; w_{\mathcal{E}}\right) - \textbf{H}^*\left(x; w_{\mathcal{E}^{*\prime}}\right) \right\|^2,
\end{gathered}
\label{l2-norm}
\end{equation}
where \(w_{\mathcal{E}^{*\prime}}=\{w_{\mathcal{E}^*},w_{\mathcal{E}^*_{\text{lora}}}\}\), \( \mathbf{H}_{t_i+n}(\cdot) \) and \( \mathbf{H}_{s_i}^*(\cdot) \) represent the hidden representation of the full LLM and sub-LLM at layer \( t_i + n \) and \( s_i \), respectively. \( \textbf{H} \) and \( \textbf{H}^* \) represent the final hidden representation of the original emulator \(\mathcal{E}\) and sub-emulator \(\mathcal{E}^*\). In addition to aligning intermediate representations, we also apply Kullback-Leibler (KL) divergence to align the output distributions:
\begin{equation}
    \mathcal{L}_{\text{KL}} = \text{KL}(\mathcal{M}(x;\{w_{\mathcal{E}},w_{\mathcal{A}}\})||\mathcal{M}^*(x;\{w_{\mathcal{E}^{*\prime}},w_{\mathcal{A}}\})),
\end{equation}
where \( \text{KL}(\cdot || \cdot) \) is the KL divergence between full LLM and sub-LLM's logits.

Finally, the objective of the entire alignment module is to jointly minimize the intermediate representation loss \( \mathcal{L}_{\text{rep}} \) and the KL loss \( \mathcal{L}_{\text{KL}} \). We use the following joint loss function to optimize the sub-LLM, where \( \alpha \) and \( \beta \) are balancing parameters that control the relative contributions of the two losses:
\begin{equation}
\min_{w_{\mathcal{E}^*_{\text{lora}}}} \frac{1}{|D_{\text{public}}|} \sum_{x \in D_{\text{public}}} \left( \alpha \mathcal{L}_{\text{rep}} + \beta \mathcal{L}_{\text{KL}} \right),
\label{distilloss}
\end{equation}
By fine-tuning only \( w_{\mathcal{E}^*_{\text{lora}}} \), the sub-LLM approximates the full LLM effectively, offering accurate gradients with minimal communication overhead during FFT.

\vspace{-10px}
\subsection{Resource-constrained Federated Fine-tuning}\label{FL Fine-Tuning}
Due to limited resources on clients, full-parameter training of the sub-LLM is infeasible. To reduce computation and communication costs, all sub-emulator parameters \( w_{\mathcal{E}^{*\prime}} \) are frozen, and only LoRA modules within the adapter \( \mathcal{A} \) are fine-tuned. The original adapter parameters \( w_{\mathcal{A}} \) remain fixed, while only \( w_{\mathcal{A}_{\text{lora}}} \) are updated to adapt the full LLM to downstream tasks efficiently.

During round \( t \), client \( C_k \) (\(1 \leq k \leq N\)) performs local update during round $t$. Instead of merely minimizing the local function $F_k{(\cdot)}$, the client $C_k$ employs a local solver to approximately minimize the following objective $h_k$:
\begin{equation}	\min_{w_{\mathcal{A}_{\text{lora}},k}}h_k(w^{(t)})=F_k(w^{(t)})+\frac\mu2\|w_{\mathcal{A}_{\text{lora}}}^{(t-1)}-w_{\mathcal{A}_{\text{lora}},k}^{(t)}\|^2,
    \label{Fed}
\end{equation}
where \( w^{(t)} = \{ w_{\mathcal{E}^{*\prime}}, w_{\mathcal{A}} , w_{\mathcal{A}_{\text{lora}},k}^{(t)} \} \), and \( w_{\mathcal{A}_{\text{lora}}}^{(t-1)} \) denotes the global LoRA parameters from the previous round. The proximal term, controlled by \( \mu \), encourages closeness between local and global LoRA parameters. The corresponding gradient update is:
\begin{equation}
\nabla h_k(w^{(t)})=\nabla_{w_{\mathcal{A}_{\text{lora}},k}}F_k(w^{(t)})+\mu(w_{\mathcal{A}_{\text{lora}}}^{(t-1)}-w_{\mathcal{A}_{\text{lora}},k}^{(t)}).
\label{local_update}
\end{equation}
After local training, clients upload only \( w_{\mathcal{A}_{\text{lora}},k}^{(t)} \), which the server aggregates using FedAvg~\cite{mcmahan:2016federated}.

\section{Cost-effectiveness Analysis} \label{sec:Analysis}

We analyze the computation, communication, and storage costs of FedSODA based on the decoder-only model, and similar reasoning applies to other architectures. Storage costs are shown in the Appendix \ref{storage cost}. In FedSODA, the server maintains both a full model and a sub-model for alignment. Each client trains a local copy of the sub-model and keeps updates consistent with the server. For simplicity, we make the following assumptions:
\vspace{-6px}
\begin{itemize}[leftmargin=2em]
    \item All clients initialize from the same pre-NF4 quantized sub-model and use identical hyperparameters (e.g., learning rate, batch size, sequence length), training for the same number of epochs.
    \item We focus on a specific client (denoted as \(k\)) to isolate local training effects from external factors like dataset size \(|D_k|\) or hardware specifications, allowing a more precise evaluation of local training’s impact on resource usage and global model performance.
\end{itemize}
\vspace{-6px}

We consider a decoder-only transformer model characterized by the following parameters: \(V\) for the vocabulary size, \(S\) for the sequence length, \(d_{\mathrm{model}}\) for the hidden dimensionality of each layer, \(d_{\mathrm{ff}}\) for the intermediate dimensionality in the feed-forward module (typically defined as \(d_{\mathrm{ff}} = f \cdot d_{\mathrm{model}}\), where \(f\) is the expansion factor), \(r\) for the LoRA rank (with \(r \ll d_{\mathrm{model}}\)), and \(m\) for the total number of transformer layers (T-layers) in the model. To distinguish different subsets of layers, we use the following definitions: \(m^G\) denotes the number of T-layers in the \textit{full model}, \(m^L\) refers to the number of T-layers in the \textit{sub-model}, \(m^A\) represents the number of T-layers of the \textit{adapter} of the sub-model, \(m^E\) represents the number of T-layers of the \textit{sub-emulator} in the sub-model, and \(m^S\) represents the number of T-layers of the sub-emulator trained during the alignment phase.

\vspace{-5px}
\subsection{Computational Cost}
Building on the notations and drawing from previous studies \cite{attention,gqa}, the computational cost of a LLM is commonly categorized into the following components:
\vspace{-6px}
\begin{itemize}[leftmargin=2em]
    \item \textbf{Embedding}:$\mathcal{O}((V + S) d_{\mathrm{model}})$;
    \item \textbf{Multi-head  Self-Attention (MHA)}:$\mathcal{O}(m(Sd_{\mathrm{model}}^2+S^2d_{\mathrm{model}}))$;
    \item \textbf{Feed-Forward Network(FFN)}: $\mathcal{O}(m Sd_{\mathrm{ff}}d_{\mathrm{model}})=\mathcal{O}(mSd_{\mathrm{model}}^2)$;
    \item \textbf{Add\&Norm}:$\mathcal{O}(mSd_{\mathrm{model}})$.
\end{itemize}
\vspace{-6px}

By incorporating QLoRA \cite{qlora}, the computational cost is $\mathcal{O}(mSrd_{\mathrm{model}})$. Hence, the forward propagation cost for the $m^f$ layers is approximated by $\mathcal{O}\bigl(m^f\bigl(Sd_{\mathrm{model}}^2 + S^2d_{\mathrm{model}} + Srd_{\mathrm{model}}\bigr)\bigr)$, and the backpropagation cost for $m^b$ layers is of a similar scale. In general, the total computation is dominated by 
$\mathcal{O}\bigl((m^f+m^b)(Sd_{\mathrm{model}}^2 + S^2d_{\mathrm{model}})\bigr)$, given $S < d_{\mathrm{model}}$ and $(V + S)d_{\mathrm{model}} \ll (m^f + m^b)(S d_{\mathrm{model}}^2 + S^2d_{\mathrm{model}})$.

Under conventional training, both forward and backward passes traverse the full model, i.e., $m^f = m^b = m^G$, leading to substantial computational cost. In contrast, FedSODA updates only the adapter during federated fine-tuning (FFT), resulting in $m^f = m^L$ and $m^b = m^A$. This significantly reduces client-side computation and enhances deployability and scalability in resource-constrained environments.

\vspace{-5px}
\subsection{Communication Cost}
Following studies \cite{attention}, we list the space complexity of the parameter size of different components in a LLM as follows.
\vspace{-6px}
\begin{itemize}[leftmargin=2em]
    \item \textbf{Embedding:} $\mathcal{O}( S d_{model})$;
    \item \textbf{T-layers:} $\mathcal{O}(md_{model}^2)$;
    \item \textbf{Output:} $\mathcal{O}( Vd_{model})$.
\end{itemize}
\vspace{-6px}

If all parameters are transmitted per FFT iteration, the communication complexity becomes 
$\mathcal{O}(Sd_{\mathrm{model}} + m^bd_{\mathrm{model}}^2 + Vd_{\mathrm{model}})$. In contrast, FedSODA transmits only the LoRA parameters for a subset of layers. 
Due to the low rank ($r \ll d_{\mathrm{model}}$) and additional quantization, the communication cost is reduced 
to $\mathcal{O}((m^S+m^A)rd_{\mathrm{model}})$ if alignment is included, or $\mathcal{O}(m^Srd_{\mathrm{model}})$ 
if not. Empirically, the overhead of transmitting quantized intermediate values is negligible compared 
to LoRA parameters, so we omit them from the analysis.

\section{Experiments} \label{sec:Experiments}
\subsection{Experimental Setup}

\subsubsection{Models and Datasets}
\begin{table}[htbp]
    \centering   
    \caption{Details of datasets being evaluated. Math: mathematical reasoning. CS: commonsense reasoning.}  
    \vspace{5px}
    \begin{tabularx}{\linewidth}{*{5}{>{\centering\arraybackslash}X}}
       \Xhline{1.5pt}
        Dataset& Domain & \# train & \# test &Answer  \\
       \hline
        GSM8K & MATH & 7473 & 1319 & Number \\
        SST-2 & CS & 67349 & 872 &Option\\
        BoolQ &CS &9427& 3270 &Option \\
        \Xhline{1.5pt}
    \end{tabularx}
    \label{dataset}
    \end{table}
    \vspace{-8px}
\begin{table*}[t]
    \centering
    \caption{IID experimental results on LLaMA3-8B, Qwen2-7B, and Mistral-7B. The results are averaged from three seeds to produce solid results. The evaluation metric is accuracy. The best result for the same setting is marked in \textbf{bold}, and \underline{underline} indicates second best.}
      \renewcommand{\arraystretch}{1.1} 
    \resizebox{\textwidth}{!}{
    \begin{tabular}{cl|ccc|cccccc}
        
        \Xhline{1.5pt}

        Model & Method & Pruning & Pre-align & Realign & \makecell[c]{GSM8K\\(4-shot)} & SST-2 & BoolQ & \makecell[c]{\# Tr.\\Params} & \makecell[c]{\# Commu.\\Params(/round)} & \makecell[c]{\# Model\\Size} \\
        
        \hline

        \multirow{4}*{LLaMA3-8B} 
         & FedPETuning &&& & \textbf{64.3} & \textbf{89.4} & \underline{68.4} & 15.2M & 30.4M & 15.0GB \\
         & FedOT & \ding{51} & \ding{51} && 60.7 & 86.3 & 66.8 & 1.3M & \textbf{2.6M} & \underline{10.1GB}  \\
         & FedBiOT & \ding{51} & \ding{51} & \ding{51} & 63.2 & \underline{88.2} & 68.1 & 1.3M & 10.2M & \underline{10.1GB} \\
         & \textbf{FedSODA} & \ding{51} & \ding{51} & \ding{51} & \underline{63.5} & 87.8 & \textbf{68.6} & 1.3M & \underline{3M} & \textbf{3.8GB} \\
        \hline
        \multirow{4}*{Qwen2-7B}
         & FedPETuning &&&& \textbf{69.5} & \underline{94.6} & \textbf{75.7} & 15.5M & 30.9M & 14.2GB \\
         & FedOT & \ding{51} & \ding{51} && 67.1 & 89.4 & 72.6 & 1.4M & \textbf{2.8M} & \underline{9.0GB} \\
         & FedBiOT & \ding{51} & \ding{51} & \ding{51} & 68.5 & \textbf{94.8} & \underline{75.1} & 1.4M & 11.1M & \underline{9.0GB} \\
         & \textbf{FedSODA} & \ding{51} & \ding{51} & \ding{51} & \underline{68.8} & 94.3 & 74.8 & 1.4M & \underline{3.3M} & \textbf{3.4GB} \\ 
        
        \hline

        \multirow{4}*{Mistral-7B} 
         & FedPETuning &&& & \textbf{58.3} & \textbf{77.5} & \underline{66.9} & 15.2M & 30.4M & 13.5GB \\
         & FedOT & \ding{51} & \ding{51} && 54.3 & 74.5 & 64.6 & 1.3M & \textbf{2.6M} & \underline{8.6GB} \\
         & FedBiOT & \ding{51} & \ding{51} & \ding{51} & \underline{57.7} & 75.7 & 66.4 & 1.3M & 10.2M & \underline{8.6GB} \\
         & \textbf{FedSODA} & \ding{51} & \ding{51} & \ding{51} & 57.2 & \underline{76.6} & \textbf{67.2} & 1.3M & \underline{3M} & \textbf{3.2GB} \\ 
        \Xhline{1.5pt}
    \end{tabular}
    }
    \label{iid}
    \vspace{-5px}
\end{table*}

In this experiment, we use three open-source LLMs: LLaMA3-8B \cite{touvron2024llama3}, Qwen2-7B \cite{baichuan2024qwen2}, and Mistral-7B \cite{mistral2023mistral7b}. All experiments were carried out in a PyTorch 2.5 environment using NVIDIA H800 GPUs. We evaluate the models on both the Natural Language Understanding (NLU) and Natural Language Generation (NLG) tasks.

For NLU, we evaluated model performance using commonsense reasoning datasets, including SST-2 \cite{SST2}, and BoolQ \cite{boolq}. In the NLG tasks, we use the GSM8K \cite{gsm8k} dataset for mathematical reasoning. The evaluation metric for these tasks is accuracy. Additionally, we use WikiText-2 \cite{merity2016pointer} as the public dataset for SGP and ODA. Table \ref{dataset} provides the statistical details of the evaluated datasets, and a comprehensive description of the public datasets and the evaluated datasets can be found in Appendix \ref{datasets}.

 \vspace{-5px}   
\subsubsection{Implementation and Hyperparameters}
\vspace{-2px}
In this experiment, we use the last three T-layers as the adapter. The pruning configuration considers a pruning factor \( n = 3 \) for each group of layers, with \( p = 4 \) pruned groups. Before starting federated fine-tuning (FFT), we applied NF4 quantization to the sub-LLM and introduced LoRA to the retained layers of the pruned groups in the sub-emulator and the adapter. The rank $r$ is set to 8, and the alpha parameter is set to 16. The sub-LLM is distilled on the server for one round to obtain the aligned sub-LLM. During FFT, the server performs alignment every 5 rounds. We employ AdamW \cite{adamw} as the optimizer, which is used to solve Equations (\ref{distilloss}) and (\ref{Fed}) on the server (sub-emulator) and client (adapter) sides, respectively. We search for the optimal learning rate from the set \( \{1 \times 10^{-5}, 5 \times 10^{-5}, 1 \times 10^{-4}\} \) and set the momentum to \( (0.9, 0.95) \). For other optimizer-related hyperparameters, we use the default settings. The results presented in this paper are based on the best combination of hyperparameters. All pre-trained LLMs and the QLoRA method used in the experiments are obtained from Hugging Face\footnote{https://huggingface.co/}. To mitigate randomness, we run the experiment with three different random seeds and report the average results. 

\vspace{-10px}
\subsubsection{Baselines}
We compare FedSODA with three baseline methods. To ensure fairness, FedOT, FedBiOT, and our approach all use the same number of T-layers as both emulator and adapter. Specifically: 1) FedPETuning \cite{fedpetuning}: LoRA fine-tuning of the full LLM. 2) FedOT \cite{federatedscope}: An extension of Offsite-tuning in the FL setting that trains without the full model, leveraging LoRA for fine-tuning. For FedOT, we use the first T-layer and the last two T-layers as the adapter, while pruning 12 T-layers from the emulator to form a sub-emulator. 3) FedBiOT \cite{fedbiot}: An enhanced version of FedOT that incorporates sub-LLM alignment in each round of FFT. For FedBiOT, the last three T-layers are used as the adapter, with the emulator configuration identical to that of FedOT.

\vspace{-10px}
\subsection{Overall Comparisons}

We first evaluate four methods in an independent and identically
distributed (IID) data scenario. Table \ref{iid} compares our method, FedSODA, with three baseline methods in fine-tuning LLaMA3-8B, Qwen2-7B, and Mistral-7B across three text datasets. Our key observations are as follows:
1) All LLMs fine-tuned with FedSODA outperform FedOT and achieve performance comparable to FedPETuning, which employs full LLMs fine-tuning. Moreover, the computational cost and communication overhead of FedSODA are on average only \textbf{8.55\%} and \textbf{9.87\%} of those of FedPETuning, demonstrating significant efficiency advantages.
2) FedOT shows a noticeable performance gap compared to FedPETuning, which may stem from the lack of sub-LLM alignment and the accumulation of gradient errors during FFT.
3) FedSODA can achieve performance comparable to FedOT with fewer communication rounds and perform similarly to FedBiOT. Although FedBiOT achieves comparable performance to FedSODA, the latter reduces communication overhead by an average of \textbf{70.6\%} compared to FedBiOT and demonstrates superior performance across most datasets.
4) Due to the adoption of NF4 quantization, our model requires only 4GB of storage, which is \textbf{24\%-26\%} of the storage required by FedPETuning. This allows sub-LLMs to be deployed on clients with limited memory, in contrast to FedOT and FedBiOT, which are unable to achieve this.

\begin{figure*}[htbp]
    \centering
    \subfloat[GSM8K Non-IID on LLaMA3-8B]{
        \centering
\includegraphics[width=0.3\linewidth,height=2.8cm]{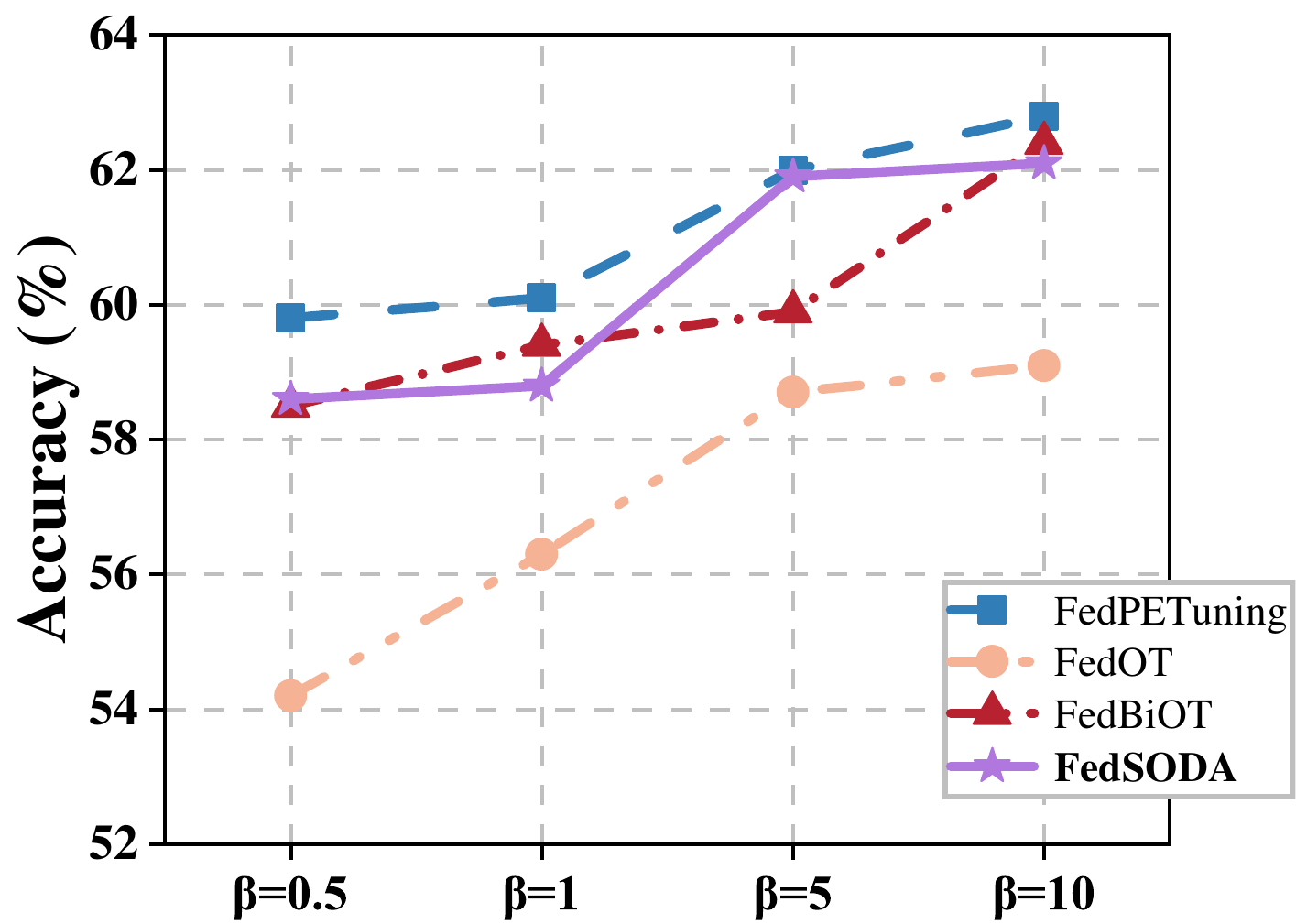}   
    }\hfill
    \vspace{-10px}
    \subfloat[GSM8K Non-IID on Qwen2-7B]{
        \centering
\includegraphics[width=0.3\linewidth,height=2.8cm]{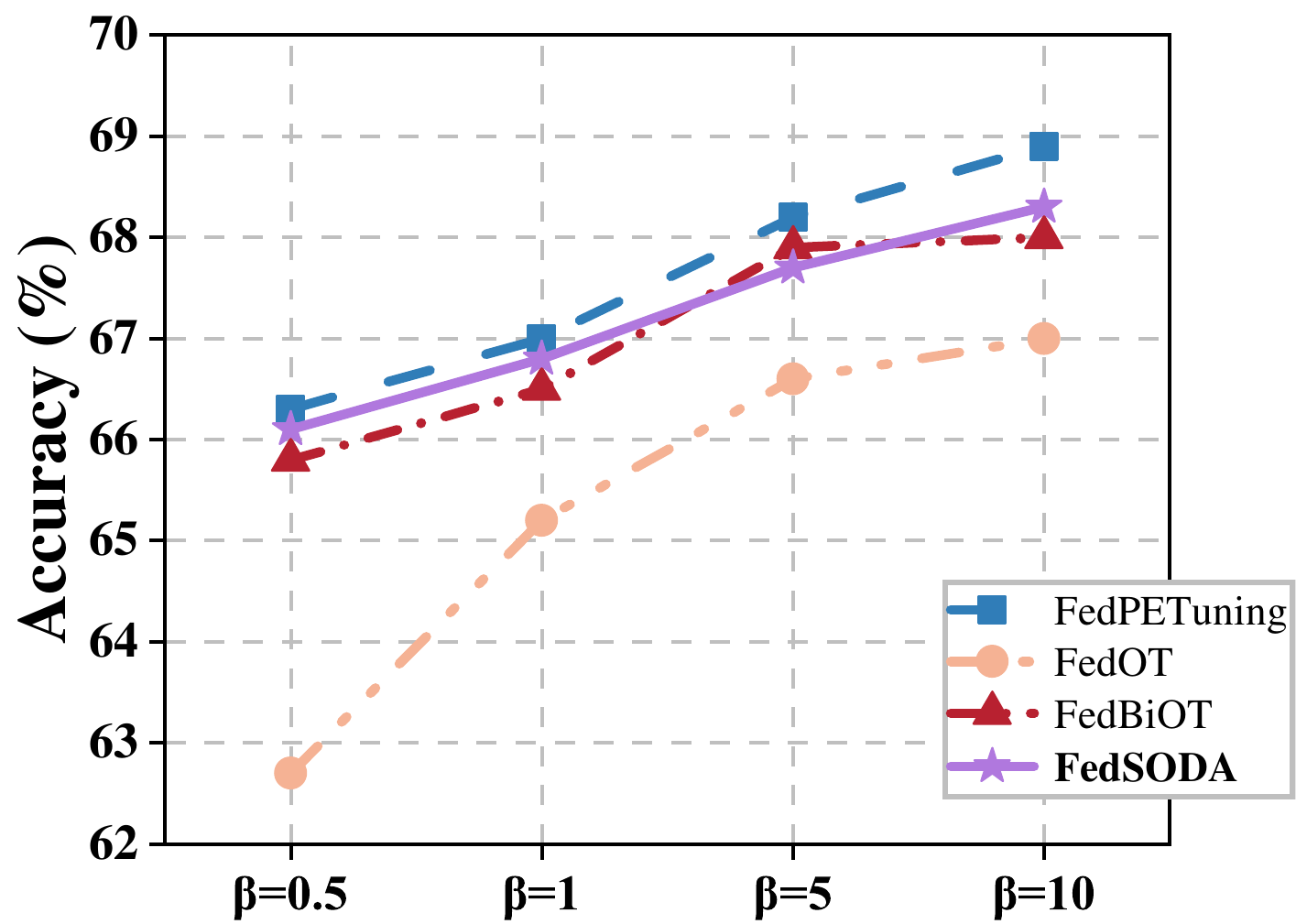}    
    }\hfill     
    \subfloat[GSM8K Non-IID on Mistral-7B]{
        \centering  
\includegraphics[width=0.3\linewidth,height=2.8cm]{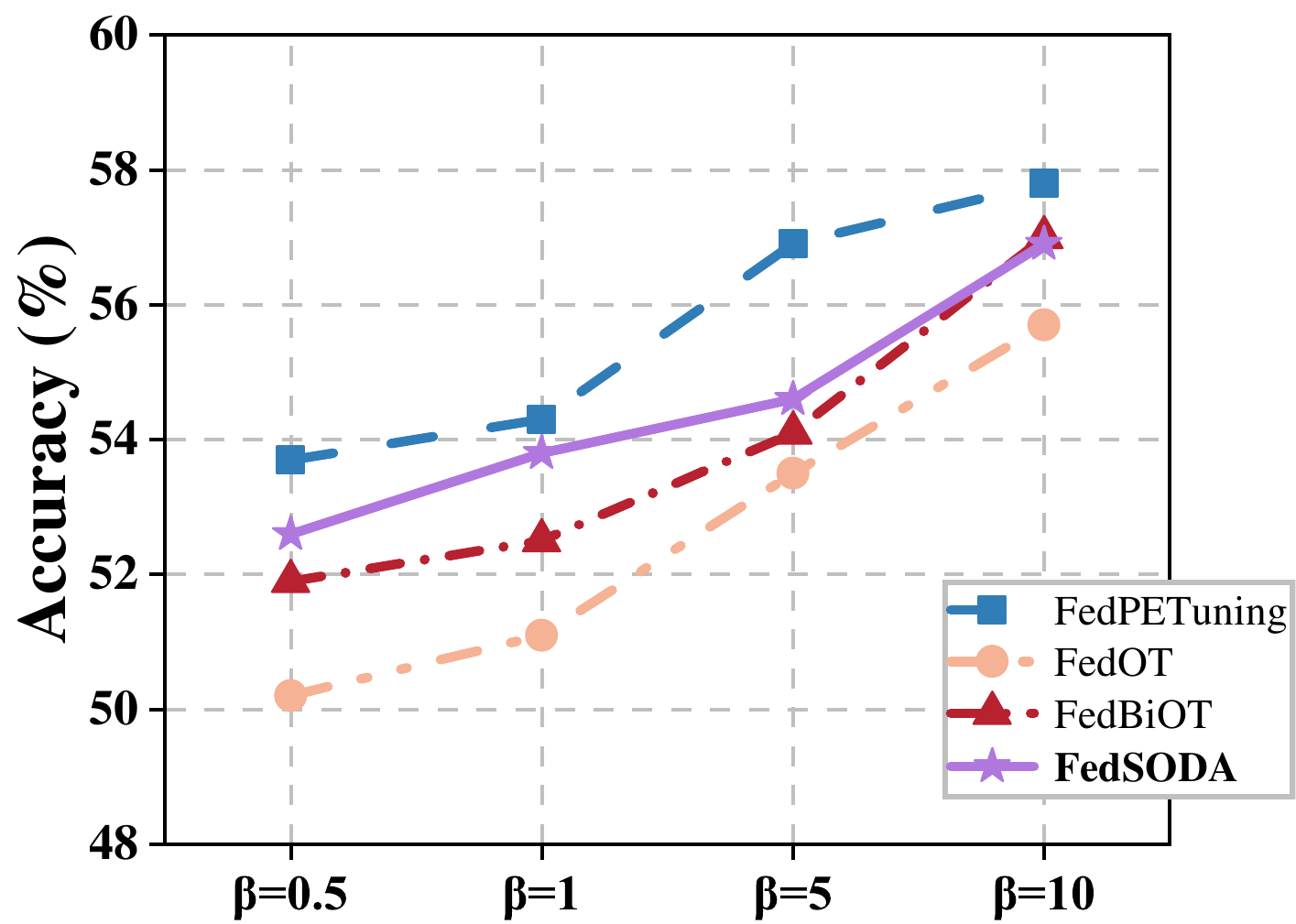}   
    }\hfill
    \subfloat[SST-2 Non-IID on LLaMA3-8B]{
        \centering
\includegraphics[width=0.3\linewidth,height=2.8cm]{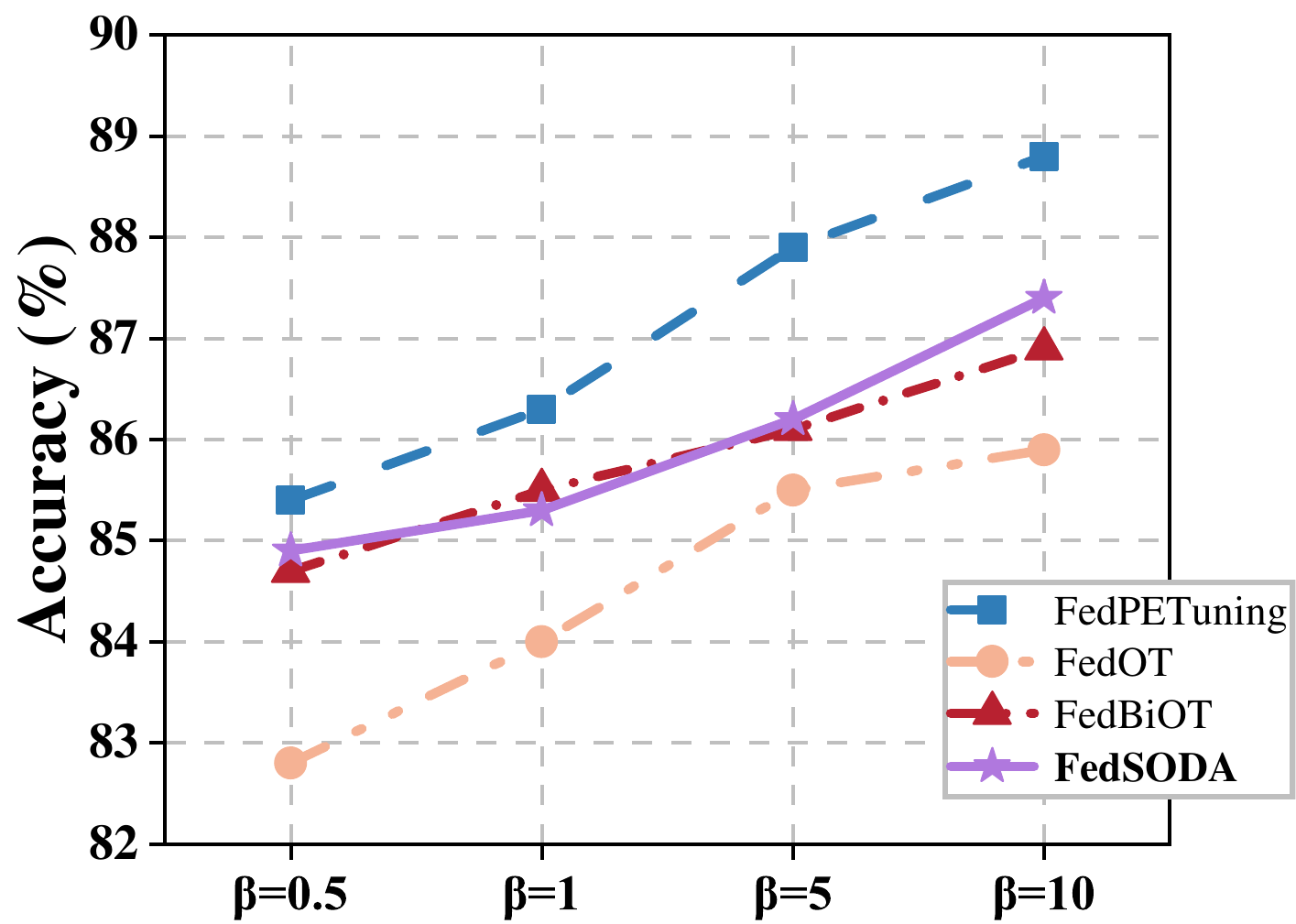}     
    }\hfill
    \subfloat[SST-2 Non-IID on Qwen2-7B]{
        \centering
\includegraphics[width=0.3\linewidth,height=2.8cm]{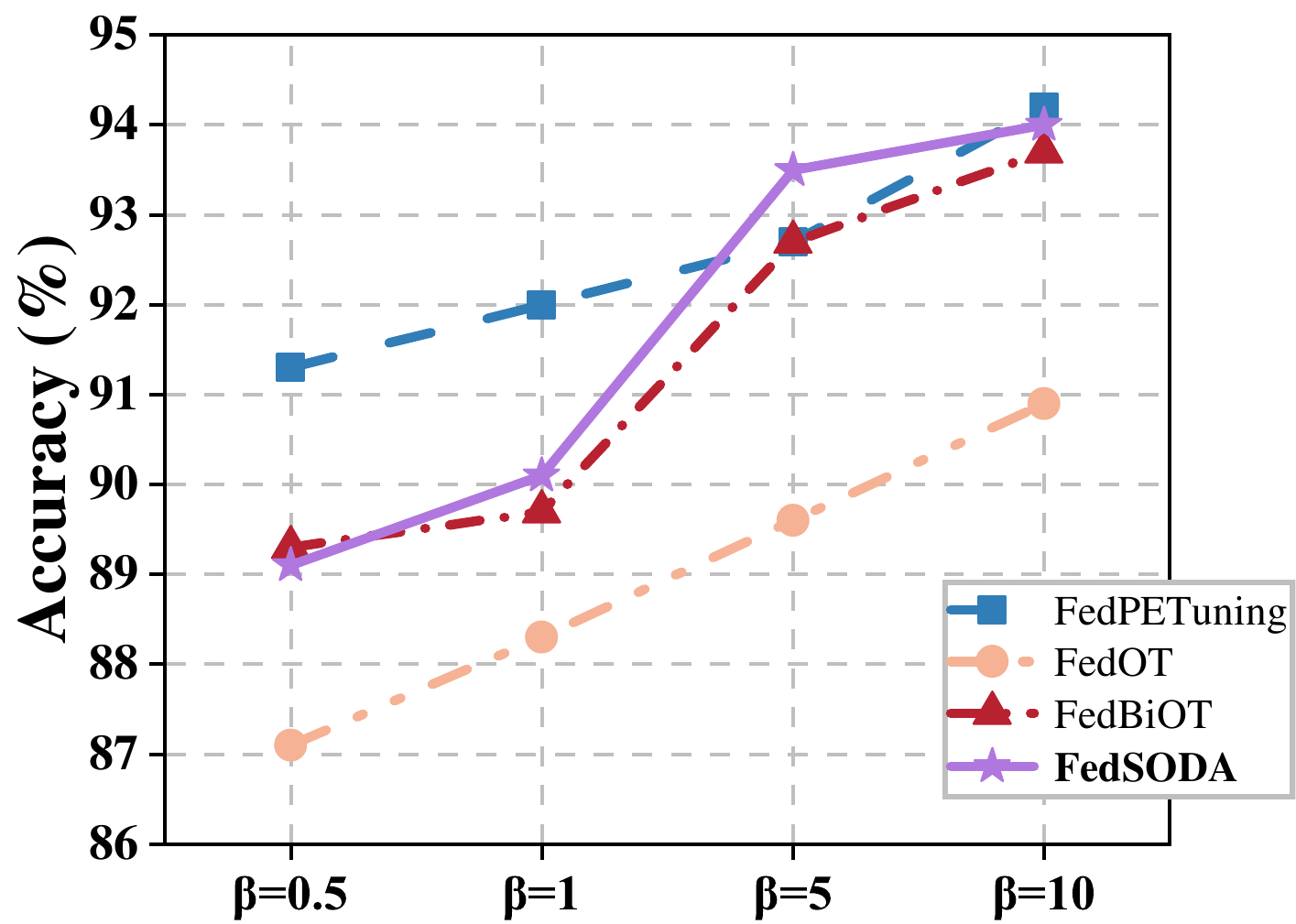}    
   }\hfill
    \subfloat[SST-2 Non-IID on Mistral-7B]{
        \centering
\includegraphics[width=0.3\linewidth,height=2.8cm]{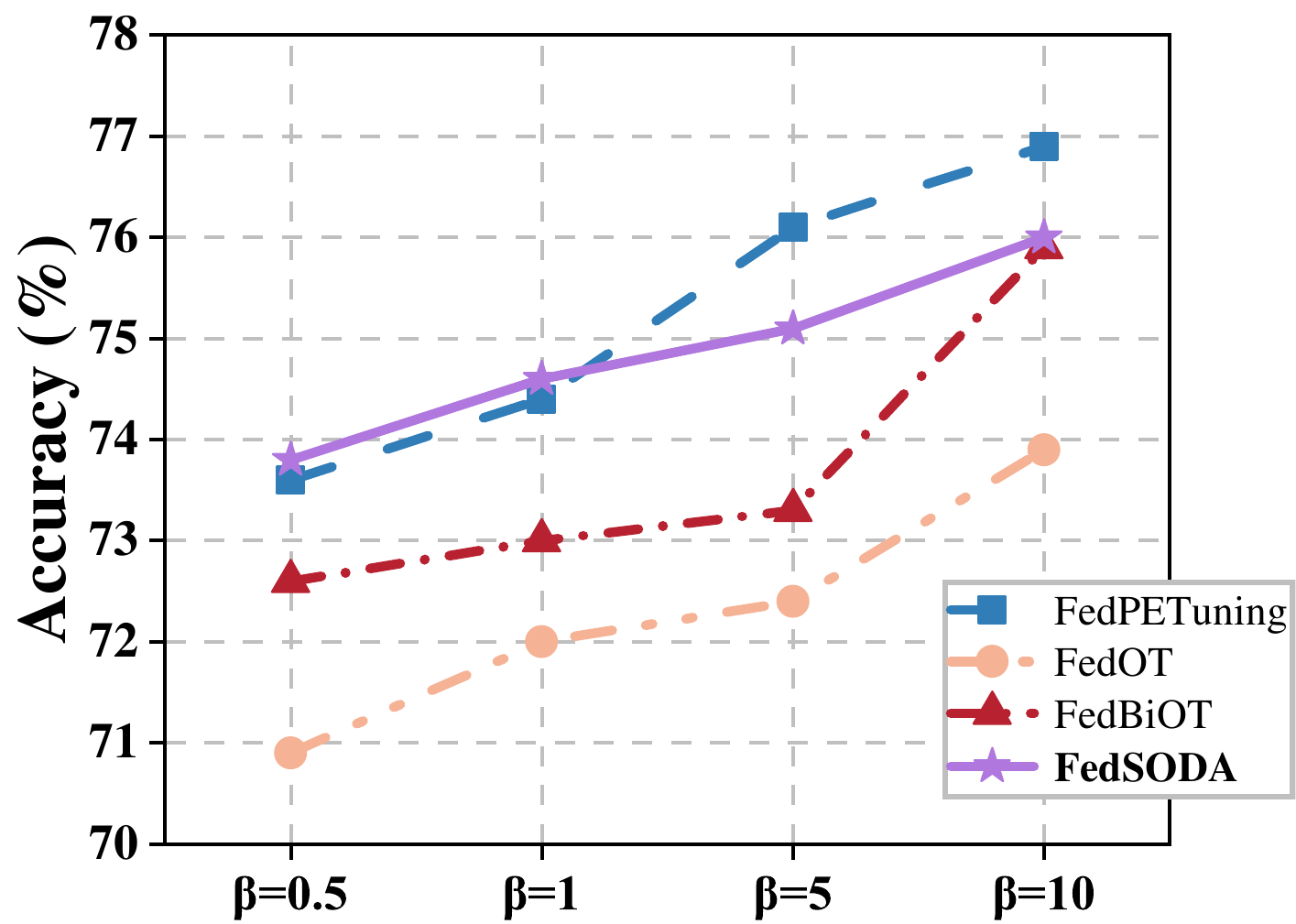}   
    }\hfill
     
    \caption{Non-IID experimental results on LLaMA3-8B, Qwen2-7B, and Mistral-7B on GSM8K and SST-2 datasets. The results are averaged from three seeds to produce solid results. The evaluation metric is accuracy.}
    \label{fig:Non_iid G_S}

\end{figure*}

\vspace{-10px}
\subsection{In-depth analysis}

\subsubsection{Impact of Data Heterogeneity}
\vspace{-3px}
In this section, we evaluate the performance of four methods in the FL scenario with Non-IID data. The dataset is partitioned into 10 clients, and a Dirichlet distribution is applied to create heterogeneous FL data scenarios. The parameter $\beta$ denotes the degree of Non-IID distribution. By adjusting the $\beta$ value, we construct three different label skewed scenarios. Figure \ref{fig:Non_iid G_S} presents the results of fine-tuning LLaMA3-8B, Qwen2-7B, and Mistral-7B using the four methods under different data heterogeneity scenarios on the GSM8K and SST-2 datasets. The Non-IID results on the BoolQ dataset are provided in the Appendix \ref{Extension}. We observe that: 
1) Under Non-IID data distribution, our method FedSODA outperforms FedOT and achieves performance comparable to that of FedPETuning. Specifically, FedSODA consistently demonstrates superior performance over FedOT under both moderate and severe Non-IID settings.
2) Moreover, it maintains competitive accuracy levels with FedPETuning across various degrees of data heterogeneity. 
3) As the degree of Non-IID increases, FedSODA exhibits strong robustness with only a slight performance drop, highlighting its effectiveness in handling real-world federated learning scenarios where data distributions are often highly skewed.
A visualization of the Non-IID data distribution is provided in Appendix \ref{Visualisations}.
\vspace{-10px}
\subsubsection{Impact of Public Dataset}
\label{impact of public dataset}
Although relying on public datasets is a widely adopted strategy in many studies, they often fall short of representing the diverse data distributions found in private datasets, which can negatively impact model performance. However, our FedSODA method can effectively alleviate this problem. The SGP module removes redundant and unimportant layers from the full LLM, enabling it to retain more general knowledge, while the ODA module can further compensate for the loss caused by pruning and alleviate gradient accumulation issues in the sub-LLM. The two modules work together to improve the robustness of the sub-LLM.

To explore the impact of the selection of public datasets on model performance, we conducted the following experiments. Specifically, we selected a subset of the GSM8K dataset (which has a significant distribution difference from WikiText-2) as the public dataset. The rest of GSM8K is used as an evaluation dataset. We then compared the performance of IID and Non-IID models under this setting to examine how the distribution difference between public and private datasets affects the final results.
\vspace{-5px}

\begin{figure}[htbp]
    \centering
    \subfloat[IID]{
        \centering
\includegraphics[width=0.43\linewidth]{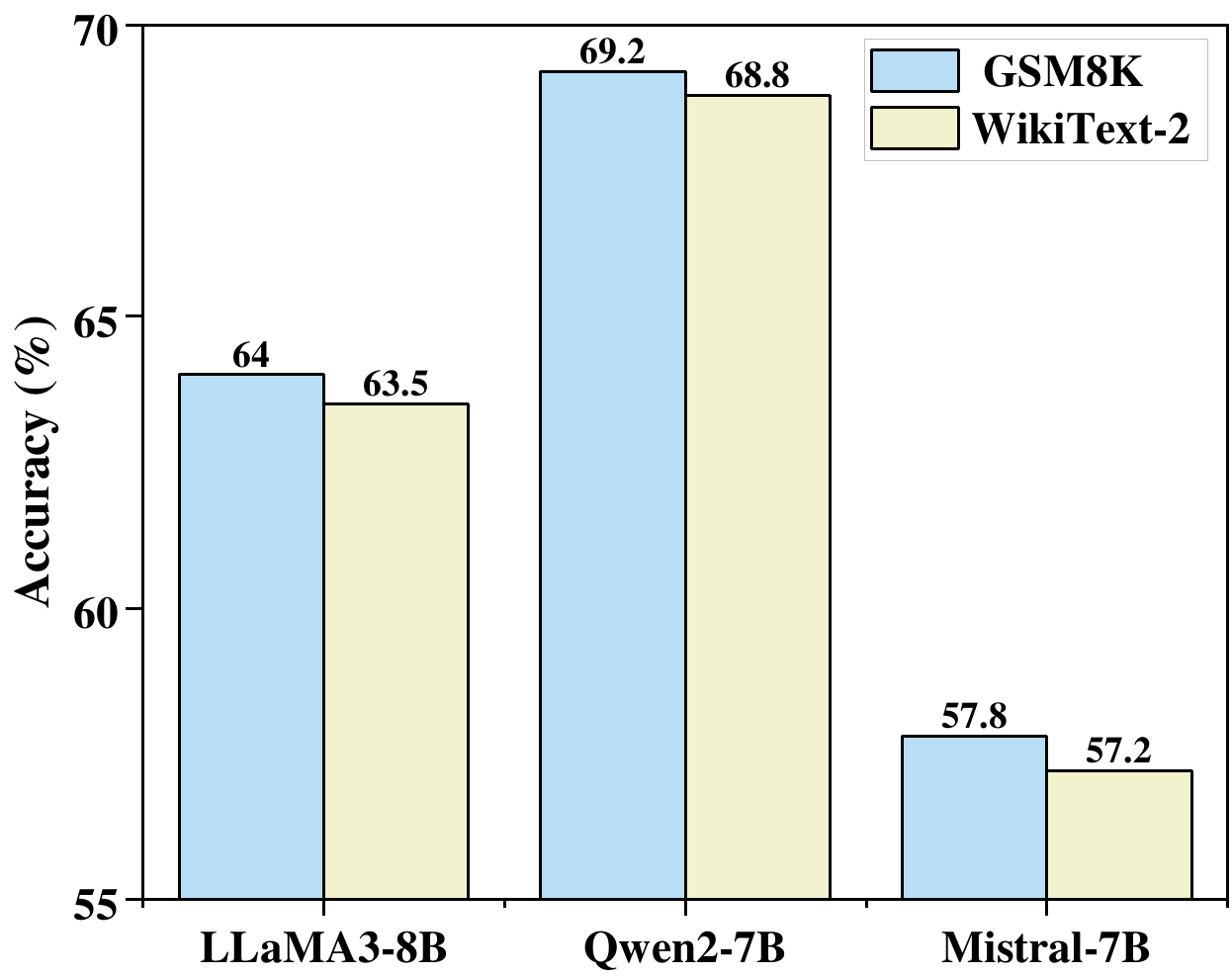}   
    }\hfill
    \vspace{-10px}
    \subfloat[Non-IID $\beta=1.0$]{
        \centering
\includegraphics[width=0.43\linewidth]{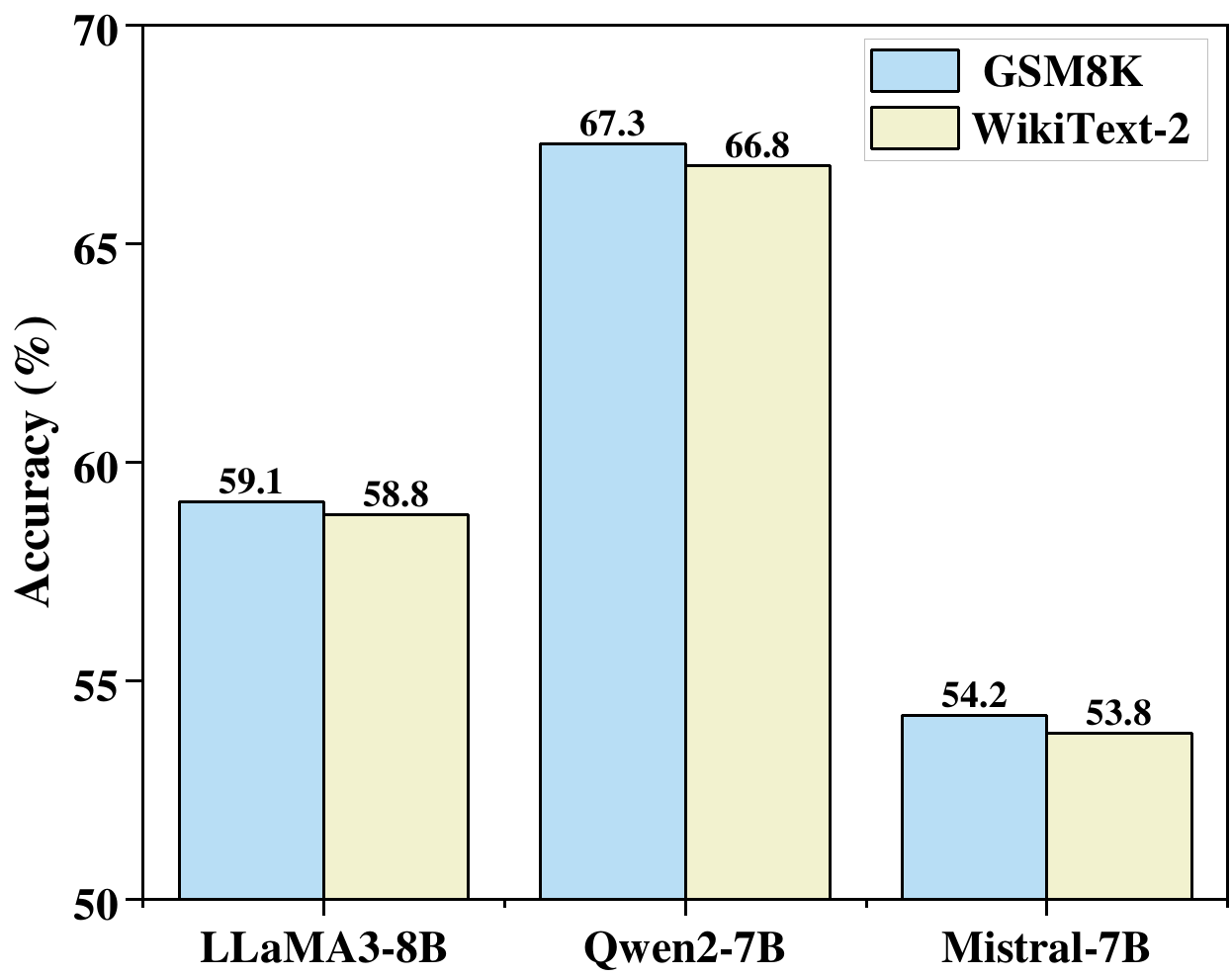}    
    }\hfill
    \vspace{10px}
    \caption{Impact of Public Dataset Selection on FedSODA Performance}    
    \label{public dataset}
\end{figure}
\vspace{10px}
The results in Figure \ref{public dataset} reveal that the choice of public dataset has a relatively limited impact on the final model performance. FedSODA performs slightly better with GSM8K as the public dataset, and shows only a minor drop when using WikiText-2 instead. This minor performance gap suggests that FedSODA is robust to the distributional mismatch between the public and private datasets. Even when the public data comes from a significantly different domain, the model still learns useful representations that generalize well to the target distribution. This property makes FedSODA more practical in real-world scenarios, where domain-specific public data may not always be available.
\vspace{-10px}

\subsection{Ablation Study}
\vspace{-5px}
\begin{table}[htbp]
    \centering
    \caption{Ablation study of FedSODA on BoolQ}
    \renewcommand{\arraystretch}{1.2}
    \setlength{\tabcolsep}{3mm}{
    \begin{tabular}{cc!{\vrule width 1pt}ccc}
        \Xhline{1.5pt}
        Pre-align&Realign &LLaMA3-8B&Qwen2-7B&Mistral-7B\\
       \hline
        \ding{55}&\ding{55}&62.4&63.9&61.3\\
        \ding{51}&\ding{55} &66.7& 70.4&65.9\\
         \ding{55}&\ding{51} &64.1&66.7&63.5\\
                  \hline
        \ding{51}&\ding{51} &\textbf{68.6}&\textbf{72.8}&\textbf{66.2}\\
        \Xhline{1.5pt}
    \end{tabular}}
    \label{Ablation study}
    \end{table}
\vspace{-5px}
To evaluate the performance of the ODA module in FedSODA, we conducted the following ablation study,
and report the accuracy in Table \ref{Ablation study}. The ablation results for GSM8K and SST-2 are provided in the Appendix \ref{Ablation}. The results demonstrate that the proposed ODA module effectively aligns the sub-LLM with the full LLM.  We observe the following: 1) Both the variant lacking alignment and the variant lacking only pre-alignment show significantly worse performance. This suggests that the lack of pre-alignment leads to a large difference between the gradients of the sub-LLM and the LLM, which hinders the convergence of fine-tuning; 2) FedSODA significantly outperforms the variants lacking realignment, emphasizing the necessity of aligning the sub-LLM during FFT.
\vspace{-10px}
\subsection{Parameter Study}
\vspace{-5px}
    \begin{table}[htbp]
    \centering
    \caption{Parameter study of FedSODA on BoolQ Fine-tuning LLaMA3-8B}
     \renewcommand{\arraystretch}{1.1}
    \setlength{\tabcolsep}{5mm}{
    \begin{tabular}{cccc}
    \Xhline{1.5pt}
         \multicolumn{2}{c}{Hyper-Parameter}&  \makecell[c]{Acc.(\%)}&\makecell[c]{Commu. Costs\\(MB/round)} \\
          \hline
        \multirow{3}*{\makecell[c]{Pruning\\Combination \\$(n, p)$}} &(2,6) & 68.4&1.51 \\
         &\textbf{(3,4)} &\textbf{68.6}& \textbf{1.43} \\
         &(4,3) &67.9 &1.39 \\
    \hline
        \multirow{3}*{\makecell[c]{Alignment\\Interval $r$}} 
         &2 & 68.1&1.69 \\
         &\textbf{5} &\textbf{68.6}&\textbf{1.43} \\
         &10 &67.2&1.35\\
    \Xhline{1.5pt}
    \end{tabular}} 
    \label{tab4}
\end{table}
\vspace{-6px}

We also conducted a study to investigate the impact of three hyperparameters in the SGP module and the ODA module: the number of layers pruned per group \( n \), the number of pruned groups \( p \), and the interval \( r \) between two alignments during FFT. In this study, we fixed the number of pruned layers to 12 and focused on exploring different combinations of \( n \) and \( p \). Table \ref{tab4} shows the impact of these three hyperparameters on the fine-tuning of the LLaMA3-8B model on the BoolQ dataset. For \( n \), we chose 3, and for \( p \), we selected 4 because it provides an effective balance between resource overhead and performance. For \( r \), longer alignment intervals may lead to the accumulation of gradient errors, while shorter intervals may hinder the sub-LLM from properly updating its weights. Additionally, shorter alignment intervals can result in higher communication overhead. The results of these three hyperparameters on the fine-tuning of the Qwen2-7B and Mistral-7B models on the BoolQ dataset can be found in the Appendix \ref{Extension}.
\vspace{-5px}
\section{Conclusions} \label{sec:Conclusions}
\vspace{-5px}
We propose a FFT framework, FedSODA, that enables clients to adapt to downstream tasks via fine-tuning without accessing the full LLM. This approach reduces computational and storage requirements on devices by compressing the full LLM and training only a small subset of parameters locally. Evaluation on multiple datasets using three open-source LLMs demonstrates that FedSODA achieves accuracy comparable to methods that fine-tune the full LLM, while significantly reducing communication overhead and storage usage.






\bibliography{mybibfile}

\begin{thebibliography}{33}
\providecommand{\natexlab}[1]{#1}
\providecommand{\url}[1]{\texttt{#1}}
\expandafter\ifx\csname urlstyle\endcsname\relax
  \providecommand{\doi}[1]{doi: #1}\else
  \providecommand{\doi}{doi: \begingroup \urlstyle{rm}\Url}\fi

\bibitem[Ainslie et~al.(2023)Ainslie, Lee-Thorp, de~Jong, Zemlyanskiy, Lebr{\'o}n, and Sanghai]{gqa}
J.~Ainslie, J.~Lee-Thorp, M.~de~Jong, Y.~Zemlyanskiy, F.~Lebr{\'o}n, and S.~Sanghai.
\newblock Gqa: Training generalized multi-query transformer models from multi-head checkpoints.
\newblock \emph{arXiv preprint arXiv:2305.13245}, 2023.

\bibitem[Bill and Eriksson(2023)]{chatrobot}
D.~Bill and T.~Eriksson.
\newblock Fine-tuning a llm using reinforcement learning from human feedback for a therapy chatbot application, 2023.

\bibitem[Che et~al.(2023)Che, Liu, Zhou, Ren, Zhou, Sheng, Dai, and Dou]{FedPepTAO}
T.~Che, J.~Liu, Y.~Zhou, J.~Ren, J.~Zhou, V.~S. Sheng, H.~Dai, and D.~Dou.
\newblock Federated learning of large language models with parameter-efficient prompt tuning and adaptive optimization.
\newblock \emph{arXiv preprint arXiv:2310.15080}, 2023.

\bibitem[Chen et~al.(2023)Chen, Tu, Li, Shen, and Chao]{chen2023}
H.-Y. Chen, C.-H. Tu, Z.~Li, H.-W. Shen, and W.-L. Chao.
\newblock On the importance and applicability of pre-training for federated learning, 2023.
\newblock URL \url{https://arxiv.org/abs/2206.11488}.

\bibitem[Cheong et~al.(2024)Cheong, Xia, Feng, Chen, and Zhang]{law1}
I.~Cheong, K.~Xia, K.~K. Feng, Q.~Z. Chen, and A.~X. Zhang.
\newblock (a) i am not a lawyer, but...: Engaging legal experts towards responsible llm policies for legal advice.
\newblock In \emph{The 2024 ACM Conference on Fairness, Accountability, and Transparency}, pages 2454--2469, 2024.

\bibitem[Clark et~al.(2019)Clark, Lee, Chang, Kwiatkowski, Collins, and Toutanova]{boolq}
C.~Clark, K.~Lee, M.-W. Chang, T.~Kwiatkowski, M.~Collins, and K.~Toutanova.
\newblock Boolq: Exploring the surprising difficulty of natural yes/no questions.
\newblock \emph{arXiv preprint arXiv:1905.10044}, 2019.

\bibitem[Cobbe et~al.(2021)Cobbe, Kosaraju, Bavarian, Chen, Jun, Kaiser, Plappert, Tworek, Hilton, Nakano, et~al.]{gsm8k}
K.~Cobbe, V.~Kosaraju, M.~Bavarian, M.~Chen, H.~Jun, L.~Kaiser, M.~Plappert, J.~Tworek, J.~Hilton, R.~Nakano, et~al.
\newblock Training verifiers to solve math word problems.
\newblock \emph{arXiv preprint arXiv:2110.14168}, 2021.

\bibitem[Cui et~al.(2023)Cui, Li, Yan, Chen, and Yuan]{law2}
J.~Cui, Z.~Li, Y.~Yan, B.~Chen, and L.~Yuan.
\newblock Chatlaw: Open-source legal large language model with integrated external knowledge bases.
\newblock \emph{arXiv preprint arXiv:2306.16092}, 2023.

\bibitem[Dettmers et~al.(2024)Dettmers, Pagnoni, Holtzman, and Zettlemoyer]{qlora}
T.~Dettmers, A.~Pagnoni, A.~Holtzman, and L.~Zettlemoyer.
\newblock Qlora: Efficient finetuning of quantized llms.
\newblock \emph{Advances in Neural Information Processing Systems}, 36, 2024.

\bibitem[Gromov et~al.(2024)Gromov, Tirumala, Shapourian, Glorioso, and Roberts]{unreasonable}
A.~Gromov, K.~Tirumala, H.~Shapourian, P.~Glorioso, and D.~A. Roberts.
\newblock The unreasonable ineffectiveness of the deeper layers.
\newblock \emph{arXiv preprint arXiv:2403.17887}, 2024.

\bibitem[Hu et~al.(2021)Hu, Shen, Wallis, Allen-Zhu, Li, Wang, Wang, and Chen]{lora}
E.~J. Hu, Y.~Shen, P.~Wallis, Z.~Allen-Zhu, Y.~Li, S.~Wang, L.~Wang, and W.~Chen.
\newblock Lora: Low-rank adaptation of large language models.
\newblock \emph{arXiv preprint arXiv:2106.09685}, 2021.

\bibitem[Kuang et~al.(2024)Kuang, Qian, Li, Chen, Gao, Pan, Xie, Li, Ding, and Zhou]{federatedscope}
W.~Kuang, B.~Qian, Z.~Li, D.~Chen, D.~Gao, X.~Pan, Y.~Xie, Y.~Li, B.~Ding, and J.~Zhou.
\newblock Federatedscope-llm: A comprehensive package for fine-tuning large language models in federated learning.
\newblock In \emph{Proceedings of the 30th ACM SIGKDD Conference on Knowledge Discovery and Data Mining}, pages 5260--5271, 2024.

\bibitem[Loshchilov and Hutter(2017)]{adamw}
I.~Loshchilov and F.~Hutter.
\newblock Decoupled weight decay regularization.
\newblock \emph{arXiv preprint arXiv:1711.05101}, 2017.

\bibitem[Mangrulkar et~al.(2022)Mangrulkar, Gugger, Debut, Belkada, Paul, and Bossan]{peft}
S.~Mangrulkar, S.~Gugger, L.~Debut, Y.~Belkada, S.~Paul, and B.~Bossan.
\newblock Peft: State-of-the-art parameter-efficient fine-tuning methods.
\newblock \url{https://github.com/huggingface/peft}, 2022.

\bibitem[McMahan et~al.(2017)McMahan, Moore, Ramage, Hampson, and y~Arcas]{mcmahan:2017communication}
B.~McMahan, E.~Moore, D.~Ramage, S.~Hampson, and B.~A. y~Arcas.
\newblock Communication-efficient learning of deep networks from decentralized data.
\newblock In \emph{Artificial intelligence and statistics}, pages 1273--1282. PMLR, 2017.

\bibitem[McMahan et~al.(2016)McMahan, Yu, Richtarik, Suresh, Bacon, et~al.]{mcmahan:2016federated}
H.~B. McMahan, F.~Yu, P.~Richtarik, A.~Suresh, D.~Bacon, et~al.
\newblock Federated learning: Strategies for improving communication efficiency.
\newblock In \emph{Proceedings of the 29th Conference on Neural Information Processing Systems (NIPS), Barcelona, Spain}, pages 5--10, 2016.

\bibitem[Merity et~al.(2016)Merity, Xiong, Bradbury, and Socher]{merity2016pointer}
S.~Merity, C.~Xiong, J.~Bradbury, and R.~Socher.
\newblock Pointer sentinel mixture models.
\newblock \emph{arXiv preprint arXiv:1609.07843}, 2016.

\bibitem[Nguyen et~al.(2022)Nguyen, Wang, Malik, Sanjabi, and Rabbat]{nguyen2022}
J.~Nguyen, J.~Wang, K.~Malik, M.~Sanjabi, and M.~Rabbat.
\newblock Where to begin? on the impact of pre-training and initialization in federated learning, 2022.
\newblock URL \url{https://arxiv.org/abs/2210.08090}.

\bibitem[Panagoulias et~al.(2024)Panagoulias, Virvou, and Tsihrintzis]{medical2}
D.~P. Panagoulias, M.~Virvou, and G.~A. Tsihrintzis.
\newblock Evaluating llm--generated multimodal diagnosis from medical images and symptom analysis.
\newblock \emph{arXiv preprint arXiv:2402.01730}, 2024.

\bibitem[Peng et~al.(2024)Peng, Fan, Chen, Wang, Pan, Wen, Zhang, and Wang]{fedpft}
Z.~Peng, X.~Fan, Y.~Chen, Z.~Wang, S.~Pan, C.~Wen, R.~Zhang, and C.~Wang.
\newblock Fedpft: Federated proxy fine-tuning of foundation models.
\newblock \emph{arXiv preprint arXiv:2404.11536}, 2024.

\bibitem[Sajjad et~al.(2023)Sajjad, Dalvi, Durrani, and Nakov]{layerdrop}
H.~Sajjad, F.~Dalvi, N.~Durrani, and P.~Nakov.
\newblock On the effect of dropping layers of pre-trained transformer models.
\newblock \emph{Computer Speech \& Language}, 77:\penalty0 101429, 2023.

\bibitem[Socher et~al.(2013)Socher, Perelygin, Wu, Chuang, Manning, Ng, and Potts]{SST2}
R.~Socher, A.~Perelygin, J.~Wu, J.~Chuang, C.~D. Manning, A.~Y. Ng, and C.~Potts.
\newblock Recursive deep models for semantic compositionality over a sentiment treebank.
\newblock In \emph{Proceedings of the 2013 conference on empirical methods in natural language processing}, pages 1631--1642, 2013.

\bibitem[Sung et~al.(2022)Sung, Cho, and Bansal]{lst}
Y.-L. Sung, J.~Cho, and M.~Bansal.
\newblock Lst: Ladder side-tuning for parameter and memory efficient transfer learning.
\newblock \emph{Advances in Neural Information Processing Systems}, 35:\penalty0 12991--13005, 2022.

\bibitem[Team(2023)]{mistral2023mistral7b}
M.~A. Team.
\newblock Mistral 7b.
\newblock \emph{arXiv preprint arXiv:2310.06825}, 2023.

\bibitem[Team(2024)]{baichuan2024qwen2}
Q.~Team.
\newblock Qwen2 technical report.
\newblock \emph{arXiv preprint arXiv:2407.10671}, 2024.

\bibitem[Touvron et~al.(2024)Touvron, Lavril, Izacard, et~al.]{touvron2024llama3}
H.~Touvron, T.~Lavril, G.~Izacard, et~al.
\newblock The llama 3 herd of models.
\newblock \emph{arXiv preprint arXiv:2407.21783}, 2024.

\bibitem[Vaswani(2017)]{attention}
A.~Vaswani.
\newblock Attention is all you need.
\newblock \emph{Advances in Neural Information Processing Systems}, 2017.

\bibitem[Wei et~al.(2024)Wei, Yang, Li, Xu, Chen, Li, Jiang, Hou, and Zhang]{medical3}
J.~Wei, D.~Yang, Y.~Li, Q.~Xu, Z.~Chen, M.~Li, Y.~Jiang, X.~Hou, and L.~Zhang.
\newblock Medaide: Towards an omni medical aide via specialized llm-based multi-agent collaboration.
\newblock \emph{arXiv preprint arXiv:2410.12532}, 2024.

\bibitem[Wu et~al.(2024)Wu, Li, Li, Ding, and Gao]{fedbiot}
F.~Wu, Z.~Li, Y.~Li, B.~Ding, and J.~Gao.
\newblock Fedbiot: Llm local fine-tuning in federated learning without full model.
\newblock In \emph{Proceedings of the 30th ACM SIGKDD Conference on Knowledge Discovery and Data Mining}, pages 3345--3355, 2024.

\bibitem[Xiao et~al.(2023)Xiao, Lin, and Han]{offsite-tuning}
G.~Xiao, J.~Lin, and S.~Han.
\newblock Offsite-tuning: Transfer learning without full model.
\newblock \emph{arXiv preprint arXiv:2302.04870}, 2023.

\bibitem[Yuan et~al.(2024)Yuan, Rastogi, Naik, Rajagopal, Goyal, Zhao, Chintagunta, and Ward]{medical1}
D.~Yuan, E.~Rastogi, G.~Naik, S.~P. Rajagopal, S.~Goyal, F.~Zhao, B.~Chintagunta, and J.~Ward.
\newblock A continued pretrained llm approach for automatic medical note generation.
\newblock \emph{arXiv preprint arXiv:2403.09057}, 2024.

\bibitem[Zhang et~al.(2023)Zhang, Yang, Dai, Wang, Yu, Qu, and Xu]{fedpetuning}
Z.~Zhang, Y.~Yang, Y.~Dai, Q.~Wang, Y.~Yu, L.~Qu, and Z.~Xu.
\newblock Fedpetuning: When federated learning meets the parameter-efficient tuning methods of pre-trained language models.
\newblock In \emph{Annual Meeting of the Association of Computational Linguistics 2023}, pages 9963--9977. Association for Computational Linguistics (ACL), 2023.

\bibitem[Zhao et~al.(2023)Zhao, Du, Li, Li, and Liu]{fedprompt}
H.~Zhao, W.~Du, F.~Li, P.~Li, and G.~Liu.
\newblock Fedprompt: Communication-efficient and privacy-preserving prompt tuning in federated learning.
\newblock In \emph{ICASSP 2023-2023 IEEE International Conference on Acoustics, Speech and Signal Processing (ICASSP)}, pages 1--5. IEEE, 2023.

\end{thebibliography}

\clearpage
\appendix
\section{Algorithm}\label{app:alo}
\begin{algorithm}[htbp]
    \caption{FedSODA}
    \label{FedSODA}
    \begin{flushleft}
        \textbf{Input}: Full LLM $\mathcal{M} = [\mathcal{E}, \mathcal{A}]$ with parameters $\Theta$,
$\mathcal{E}$ contains $L_\mathcal{E}$ transformer layers, 
pruning configuration with $n$ pruned layers per group and $p$ groups, 
public dataset $D_{\text{public}}$,
$N$ clients with local datasets $\{D_1, \dots, D_N\}$, 
federated fine-tuning for $I$ iterations, 
$H$ local update epochs, 
$E_p$ pre-alignment epochs,
$E_r$ realignment epochs,
realignment interval $r$.\\
    \textbf{Output}: Fine-tuned full LLM's parameters $\Theta$.
    \end{flushleft}
    \begin{algorithmic}[1] 
        \State $\mathcal{M}^*$, $\mathcal{I}$ $\leftarrow$ Similarity Group Pruning($\mathcal{E}$, $ L_{\mathcal{E}}$, $\mathcal{A}$, $ D_{\text{public}}$, $n$, $p$)  \Comment{See Algo. \ref{alg.Layer prune} for details}
        \State Initialize the LoRA parameters$\{w_{\mathcal{E}^*_{\text{lora}}}, w_{\mathcal{A}_{\text{lora}}}\}$ for $\mathcal{M}^*$ and perform NF4 quantization.
        \For{$e = 1,\dots,E_p$} \Comment{Pre-alignment}
        \State update $w_{\mathcal{E}^*_{\text{lora}}}$ with public dataset $D_{\text{public}}$ and $\mathcal{I}$ using Equation (\ref{distilloss}) on the server
	\EndFor
        \State send $\mathcal{M}^*$ to each client
     \For {$t=1,\dots,I$} \Comment{FFT for $I$ iterations}
		\For{each client $C_k,k \in [N]$ \textbf{in parallel}}
    \If{$t \bmod r = 0$} \Comment{Realignment}
    \For{$e = 1,\dots,E_r$}
    \State randomly select public datasets $D_{\text{public}}$ for $w_{\mathcal{E}^*_{\text{lora}}}$ alignment on the server
    \EndFor
    \State send $w_{\mathcal{E}^*_{\text{lora}}}$to all clients
    \EndIf
     \For{$h = 1, \dots, H$}\Comment{Local Updates}
        \State train local model $w_{\mathcal{A}_{\text{lora}},k}^{(t)}$ with parameters $w_{\mathcal{A}_{\text{lora}}}^{(t-1)}$ and private dataset $D_k$ using Equation (\ref{Fed})
    \EndFor
    \State upload $w_{\mathcal{A}_{\text{lora}},k}^{(t)}$ to the server and merge using FedAvg
    \State send $w_{\mathcal{A}_{\text{lora}}}^{(t)}$ to all clients
    \EndFor
    \EndFor
    \State $\Theta\leftarrow \Theta \cup w_{\mathcal{A}_{\text{lora}}}$
    \Comment{Plug \( w_{\mathcal{A}_{\text{lora}}} \) into the full LLM $\mathcal{M}$}
    \State \textbf{return} $\Theta$
    \end{algorithmic}
\end{algorithm}

\section{Discussion on the Use of Public Datasets}
\label{app:public datasets}
The public dataset $D_{\text{public}}$ used for similarity group pruning (SGP) and orchestrated distillation alignment (ODA) may not accurately reflect the distribution of downstream private datasets. We acknowledge this limitation and provide several clarifications. First, in federated learning scenarios, it is often infeasible to access downstream data due to privacy or legal constraints, making the use of $D_{\text{public}}$ a necessary compromise. Second, many well-established studies \cite{fedpft,fedbiot,federatedscope} adopt public datasets during compression and alignment phases to ensure reproducibility and comparability. Third, the pruning and alignment steps are designed to retain general capabilities rather than task-specific performance, meaning that even if there is a significant gap between the public dataset and the downstream task datasets, it is still possible to construct a sub-LLM that generalizes effectively across various downstream tasks. Finally, as demonstrated in our experiments (Section \ref{impact of public dataset}), models compressed using public datasets continue to perform well on downstream tasks, providing empirical evidence for the robustness of our approach.

\section{Storage Cost}
\label{storage cost}
NF4 quantization stores each parameter as $4$ bits instead of 32, achieving up to a $8\times$ compression ratio. Therefore, a model with $P$ parameters requires $4P$ bytes in FP32, but only $P/2$ bytes in NF4. NF4 quantization requires additional quantization constants. For example, using a block size of 64 for $W$, the quantization constant adds an average of $32/64 = 0.5$ bits  per parameter. With Double Quantization \cite{qlora}, the memory footprint can be further reduced, from $32/64 = 0.5$ bits to $8/64 + 32/(64 \times 256) = 0.127 $ bits. Although NF4 quantization necessitates extra quantization data, the overall storage is typically reduced by $6-7 \times$ compared to FP32. This property proves particularly beneficial for deploying LLMs on memory-constrained clients.
\section{Datasets}\label{datasets}
\subsection{WikiText-2}
WikiText-2 \cite{merity2016pointer} language modeling dataset is a collection of over 100 million tokens derived from the set of verified Good and Featured articles on Wikipedia. It is made available under the Creative Commons Attribution-ShareAlike License. In comparison to the preprocessed version of the Penn Treebank (PTB), WikiText-2 is more than twice as large, while WikiText-103 is over 110 times larger. The WikiText dataset also features a significantly larger vocabulary and preserves the original case, punctuation, and numbers—elements that are removed in the PTB version. Since it consists of full articles, the dataset is particularly well-suited for models that can leverage long-term dependencies.

\subsection{GSM8K}

GSM8K (Grade School Math 8K) \cite{gsm8k} is a dataset designed for solving arithmetic word problems. It contains 8792 examples of grade school-level math problems, each consisting of a question along with a corresponding passage that provides the necessary information to solve the problem. The dataset is commonly used to train and evaluate models on mathematical reasoning and problem-solving tasks. The problems in GSM8K require models to understand the problem context and apply basic arithmetic operations to find the correct answer. The dataset is split into a training set with 7473 problems and a test set with 1319 problems. GSM8K is widely used in natural language processing (NLP) research, particularly in areas related to arithmetic reasoning and logical deduction.
\subsection{SST-2}
SST-2 (Stanford Sentiment Treebank 2)~\cite{SST2} is a sentiment analysis dataset derived from movie reviews. It contains 67349 sentences labeled as either "positive" or "negative", representing the sentiment expressed in each sentence. The dataset is a subset of the larger Stanford Sentiment Treebank and is commonly used for training and evaluating models on binary sentiment classification tasks. The dataset is split into a training set with 67349 sentences and a test set with 872 sentences.
\subsection{BoolQ}
BoolQ (Boolean Questions) \cite{boolq} is a dataset designed for the task of natural language inference (NLI) and question answering. It consists of 12697 human-annotated yes/no questions, where each question is paired with a passage that provides contextual information. The dataset aims to test a model's ability to answer simple yes/no questions based on a given passage, making it a valuable resource for evaluating models on the task of reading comprehension and reasoning. The dataset is split into a training set with 9427 questions and a test set with 3270 questions.

\section{Ablation Study on Other Datasets}\label{Ablation}

\begin{table}[!]
    \centering
    
     \caption{Ablation study of FedSODA on GSM8K}
     \renewcommand{\arraystretch}{1.2}
    \setlength{\tabcolsep}{3mm}{
    \begin{tabular}{cc!{\vrule width 1pt}ccc}
         \Xhline{1.5pt}
        Pre-align&Realign& LLaMA3-8B&Qwen2-7B&Mistral-7B\\
       \hline
         \ding{55}&\ding{55} &60.3&66.4&53.9\\
         \ding{51}&\ding{55}&62.7& 67.8&56.5\\
        \ding{55}&\ding{51} &62.1&67&56.1\\
         \hline
\ding{51}&\ding{51}&\textbf{63.5}&\textbf{68.8}&\textbf{57.2}\\
         \Xhline{1.5pt}
    \end{tabular}
   }
    
    \label{gsm8k Ablation study}
    \end{table}
    \vspace{5px}
    
\begin{table}[!]
    \centering
     \caption{Ablation study of FedSODA on SST-2}
   \renewcommand{\arraystretch}{1.2}
    \setlength{\tabcolsep}{3mm}{
    \begin{tabular}{cc!{\vrule width 1pt}ccc}
         \Xhline{1.5pt}
          Pre-align&Realign &LLaMA3-8B&Qwen2-7B&Mistral-7B\\
       \hline
        \ding{55}&\ding{55} &62.4&63.9&61.3\\
        \ding{51}&\ding{55} &66.7& 70.4&65.9\\
        \ding{55}&\ding{51}&64.1&66.7&63.5\\
         \hline
\ding{51}&\ding{51}&\textbf{68.6}&\textbf{72.8}&\textbf{66.2}\\
         \Xhline{1.5pt}
    \end{tabular}}
    
    \label{sst Ablation study}
    \end{table}
    
The experiment results of ablation study on GSM8K and SST-2 are shown in Table \ref{gsm8k Ablation study} and Table \ref{sst Ablation study}. The performance of our DDP module on both datasets further validates the effectiveness of the ODA module.

\section{Extension Experimenal Results}\label{Extension}
\begin{table}[!]
    \centering
    \caption{Parameter study of FedSODA on BoolQ fine-tuning Qwen2-7B}
    \renewcommand{\arraystretch}{1.2}
    \setlength{\tabcolsep}{4.5mm}{
    \begin{tabular}{cccc}
     \Xhline{1.5pt}
         \multicolumn{2}{c}{Hyper-Parameter} & \makecell[c]{Acc.(\%)} & \makecell[c]{Commu. Costs\\(MB/round)} \\
          \hline
        \multirow{3}{*}{\makecell[c]{Pruning\\Combination \\$(n, p)$}} & (2,6) & 74.6 & 1.66 \\
         & \textbf{(3,4)} & \textbf{74.8} & \textbf{1.57} \\
         & (4,3) & 73.4 & 1.52 \\
    \hline
        \multirow{3}{*}{\makecell[c]{Alignment\\Interval $r$}} 
         & 2 & 68.1 & 1.84 \\
         & \textbf{5} & \textbf{68.6} & \textbf{1.57} \\
         & 10 & 67.4 & 1.46 \\
     \Xhline{1.5pt}
    \end{tabular} }
    \label{tab7}    
\end{table}  
\begin{table}[!]
    \centering
    \caption{Parameter study of FedSODA on BoolQ fine-tuning Mistral-7B}
    \renewcommand{\arraystretch}{1.2}
    \setlength{\tabcolsep}{4.5mm}{
    \begin{tabular}{cccc}
     \Xhline{1.5pt}
         \multicolumn{2}{c}{Hyper-Parameter}&  \makecell[c]{Acc.(\%)}&\makecell[c]{Commu. Costs\\(MB/round)} \\
          \hline
        \multirow{3}*{\makecell[c]{Pruning\\combination \\$(n, p)$}} &(2,6) & 67&1.51 \\
         &\textbf{(3,4)} &\textbf{67.2}& \textbf{1.43} \\
         &(4,3) &66.7 &1.39 \\
    \hline
        \multirow{3}*{\makecell[c]{Alignment\\interval $r$}} 
         &2 & 66.4&1.69 \\
         &\textbf{5} &\textbf{67.2}&\textbf{1.43} \\
         &10 &65.4&1.35\\
     \Xhline{1.5pt}
    \end{tabular}}
    \label{tab8}    
\end{table}

\begin{figure*}[ht]
    \centering
    \subfloat[BoolQ Non-IID on LLaMA3-8B]{
        \centering
  \includegraphics[width=0.3\linewidth]{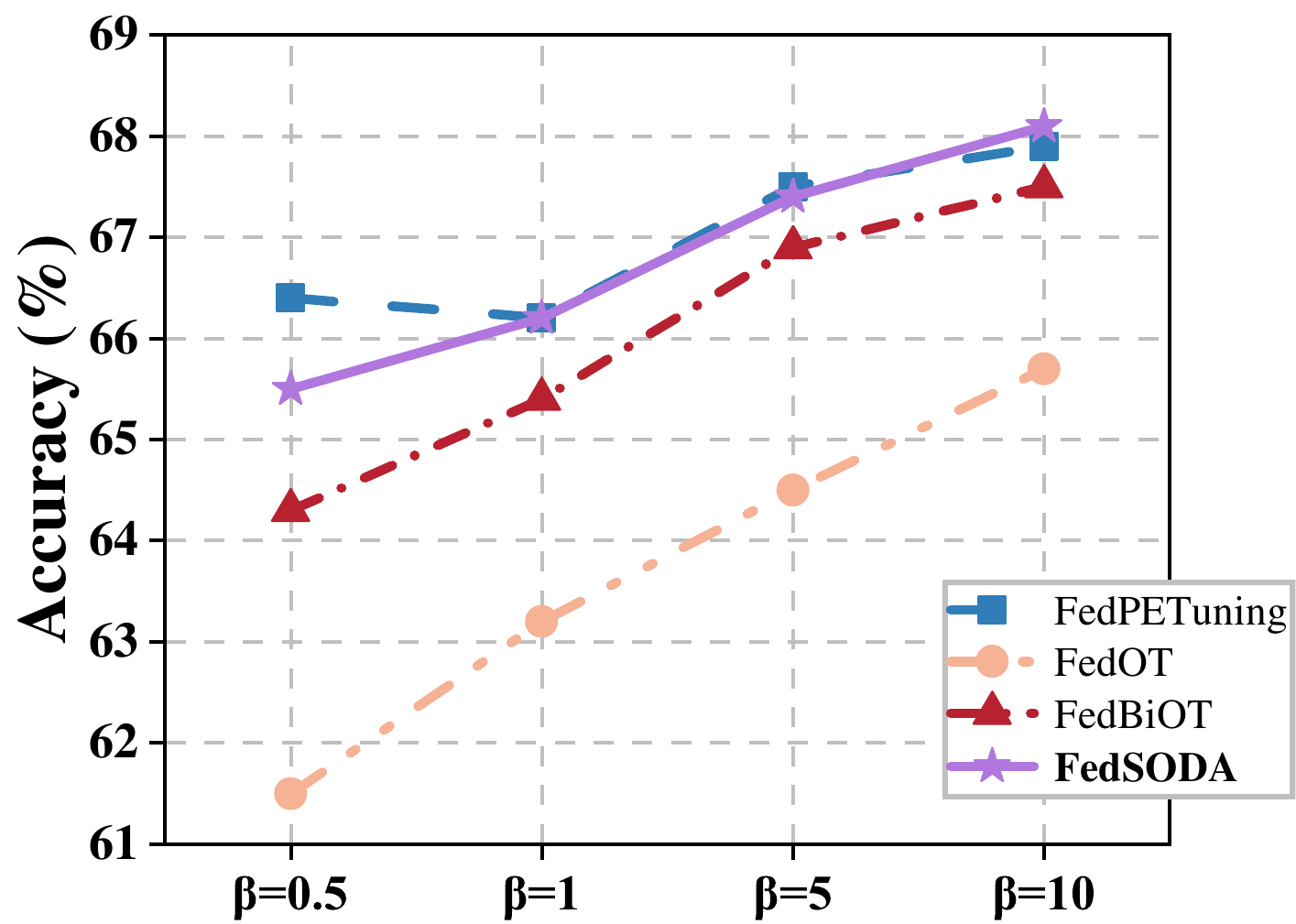}   
    }\hfill
    \subfloat[BoolQ Non-IID on Qwen2-7B]{
        \centering
    \includegraphics[width=0.3\linewidth]{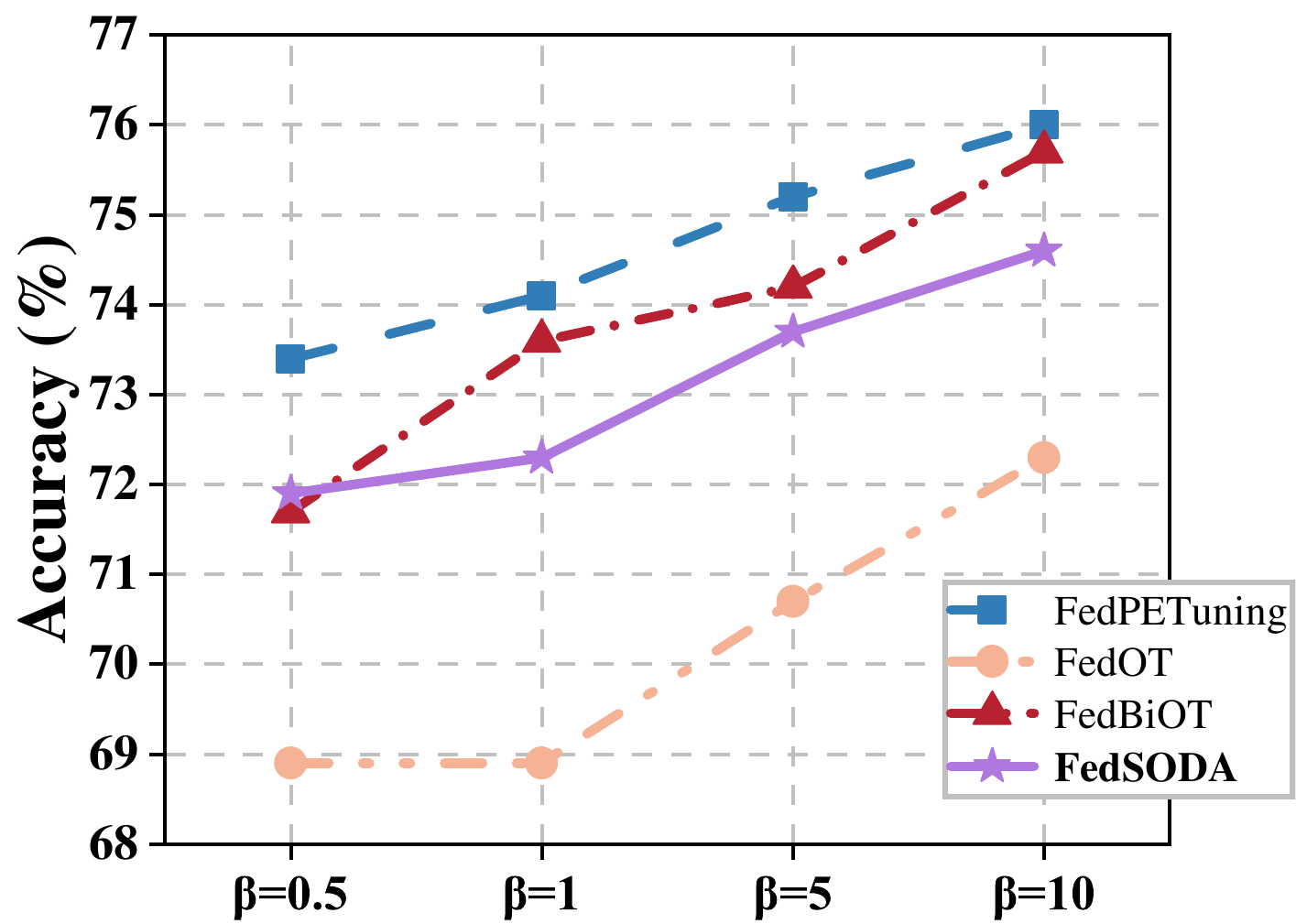}    
    }\hfill
    \subfloat[BoolQ Non-IID on Mistral-7B]{
        \centering
    \includegraphics[width=0.3\linewidth]{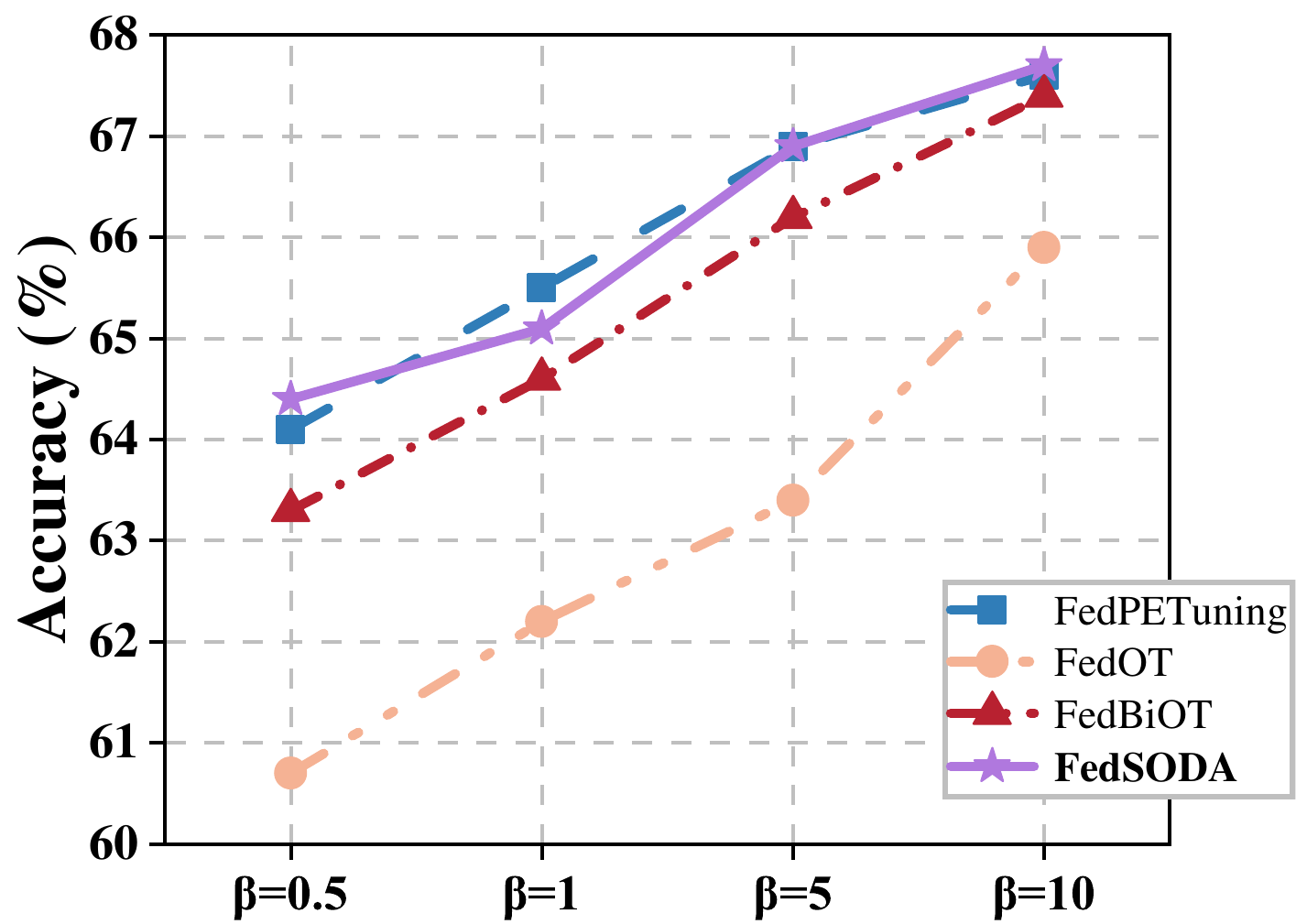}   
    }\hfill
    \vspace{5px}
    \caption{Non-IID experimental results on LLaMA3-8B, Qwen2-7B, and Mistral-7B on BoolQ dataset. The results are averaged from three seeds to produce solid results. The evaluation metric is accuracy.}
    \label{fig:Non_iid BoolQ}
\end{figure*}

\begin{table}[!]
    \centering
    \caption{Communication rounds required to achieve the same performance by all methods. We list the communication rounds required for all methods to achieve the same performance as FedOT.}
    \vspace{5px}
    \setlength{\tabcolsep}{5mm}{
    \begin{tabular}{cccc}
     \Xhline{1.5pt}
         Method& GSM8K & SST-2 &BoolQ\\
          \hline
        FedPETuning&14 & 10& 9\\
        FedOT &40 &28 &23\\
        FedBiOT &18 &13 &14\\
        \textbf{FedSODA} &\textbf{16} &\textbf{15}&\textbf{17}\\  
     \Xhline{1.5pt}
    \end{tabular}} 
    \label{tab9}
\end{table}
We extended our experiments by conducting parameter learning for Qwen2-7B and Mistral-7B as shown in Table \ref{tab7} and Table \ref{tab8}, and we performed convergence analysis on three datasets for LLaMA3-8B as shown in Table \ref{tab9}. The results of the BoolQ dataset under Non-IID are shown in Figure \ref{fig:Non_iid BoolQ}.

\section{Dirichlet Distribution Visualisations}\label{Visualisations}
\begin{figure*}[ht]
    \centering
    \subfloat[GSM8K $\beta=0.5$]{
        \includegraphics[width=0.23\textwidth]{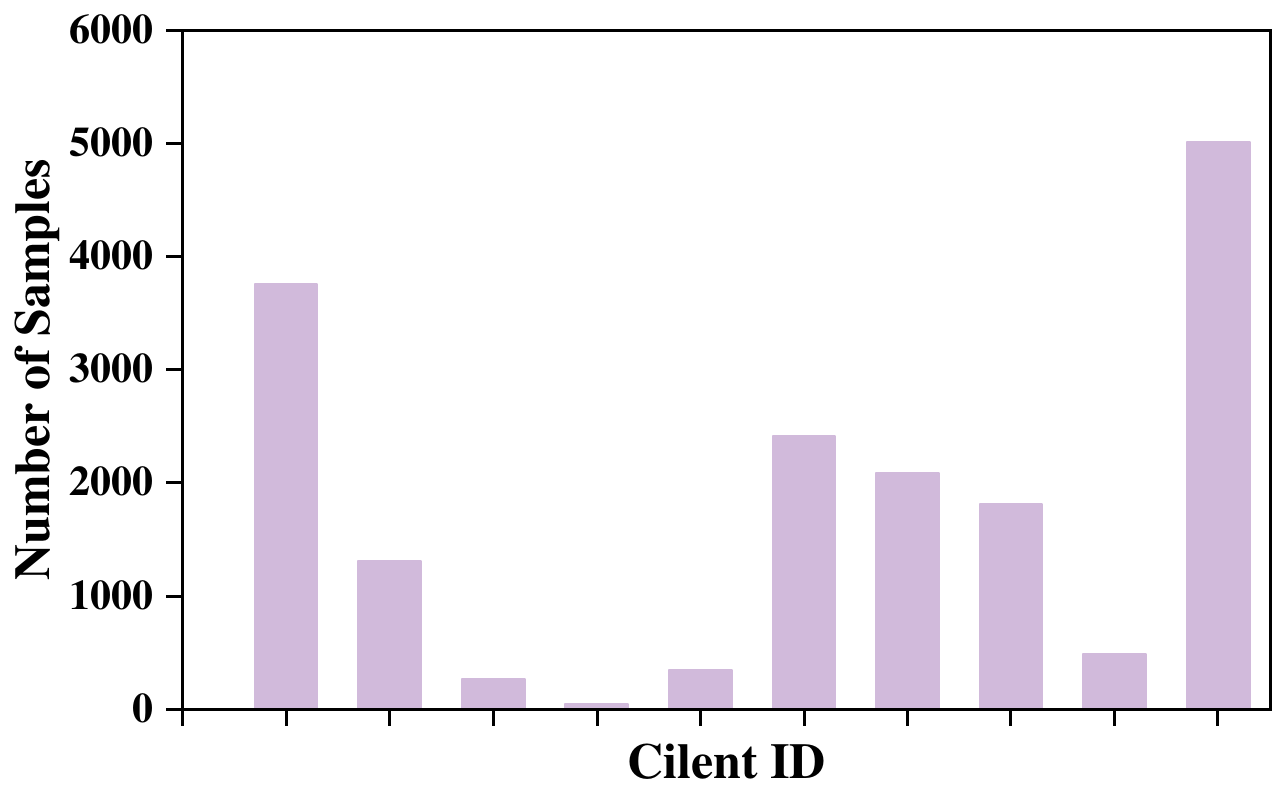}
        \label{fig:GSM8K_0.5}
    }\hfill
    \subfloat[GSM8K $\beta=1.0$]{
        \includegraphics[width=0.23\textwidth]{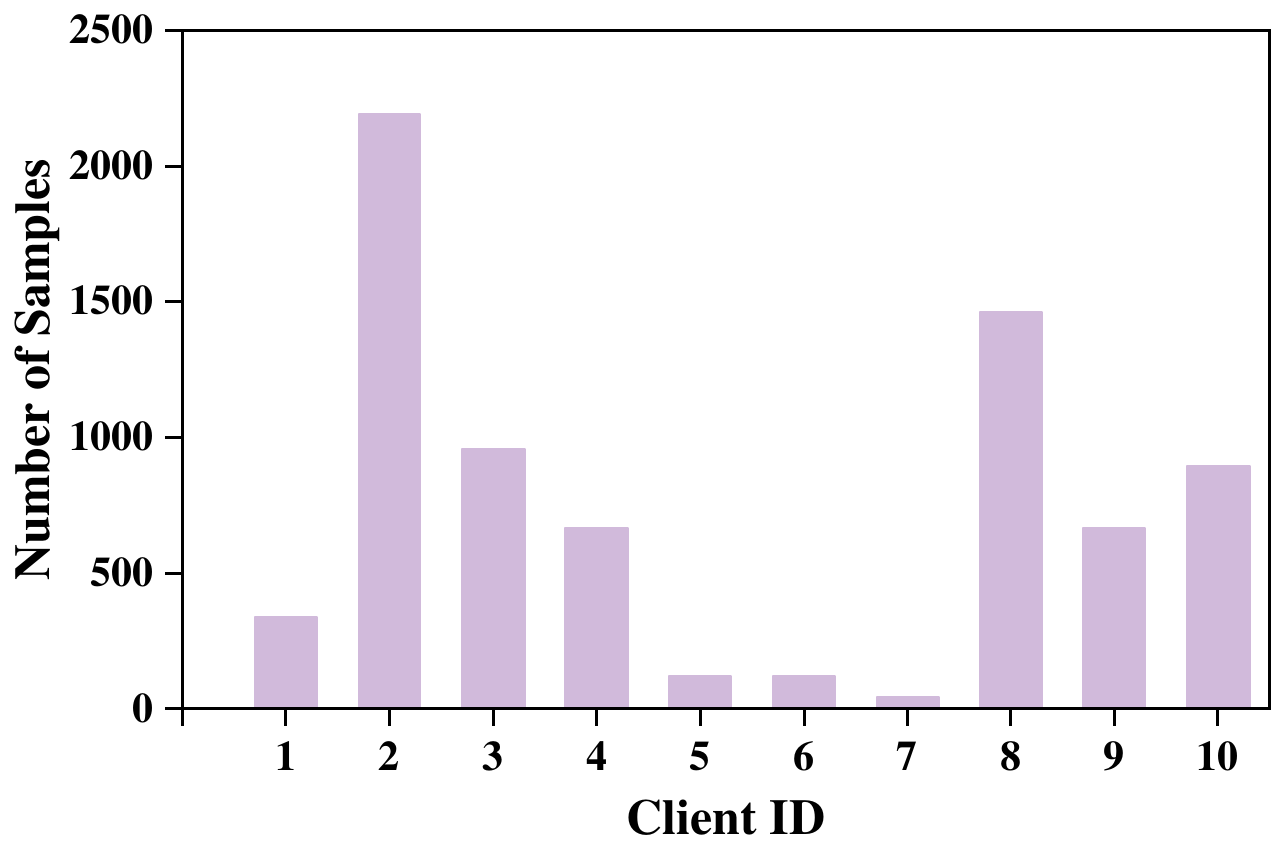}
        \label{fig:GSM8K_1.0}
    }\hfill
    \subfloat[GSM8K $\beta=5.0$]{
        \includegraphics[width=0.23\textwidth]{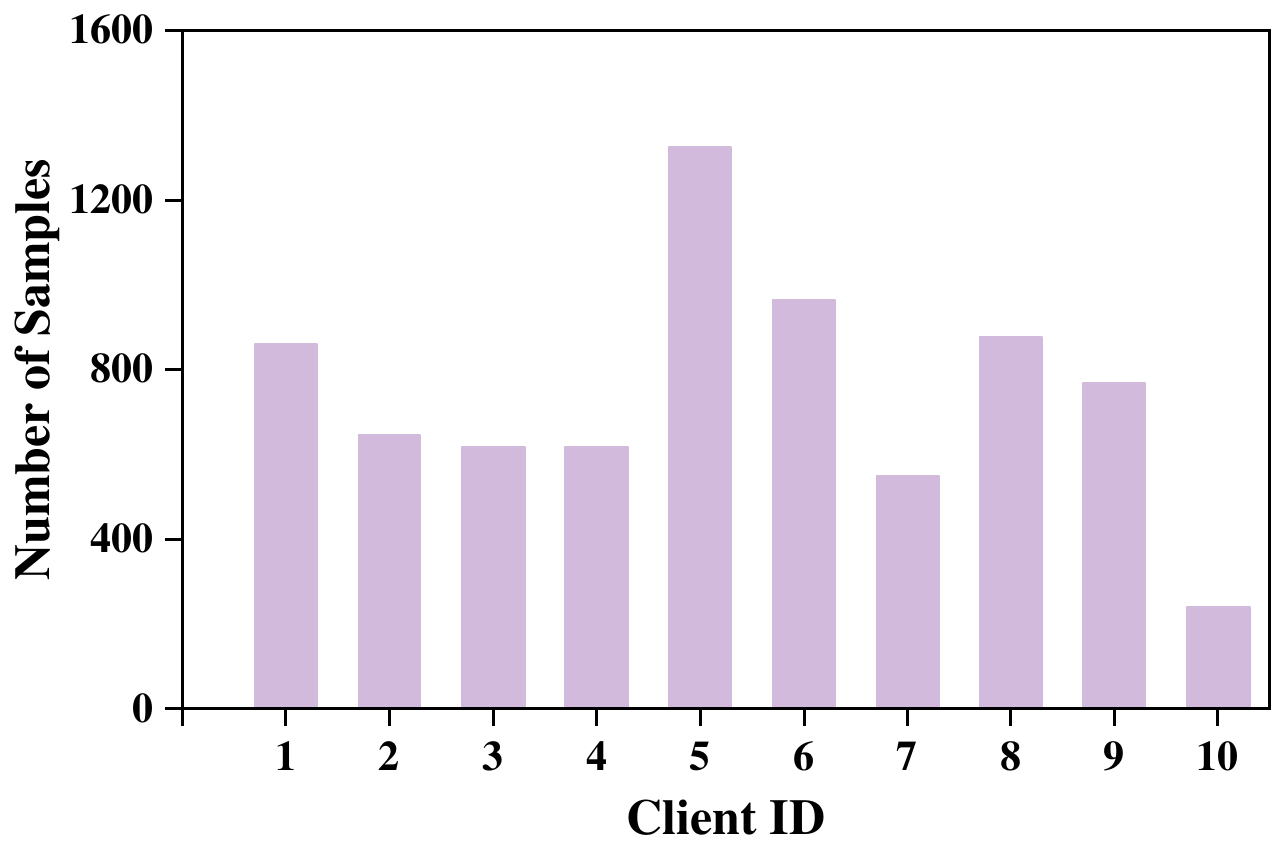}
        \label{fig:GSM8K_5.0}
    }\hfill
    \subfloat[GSM8K $\beta=10.0$]{
        \includegraphics[width=0.23\textwidth]{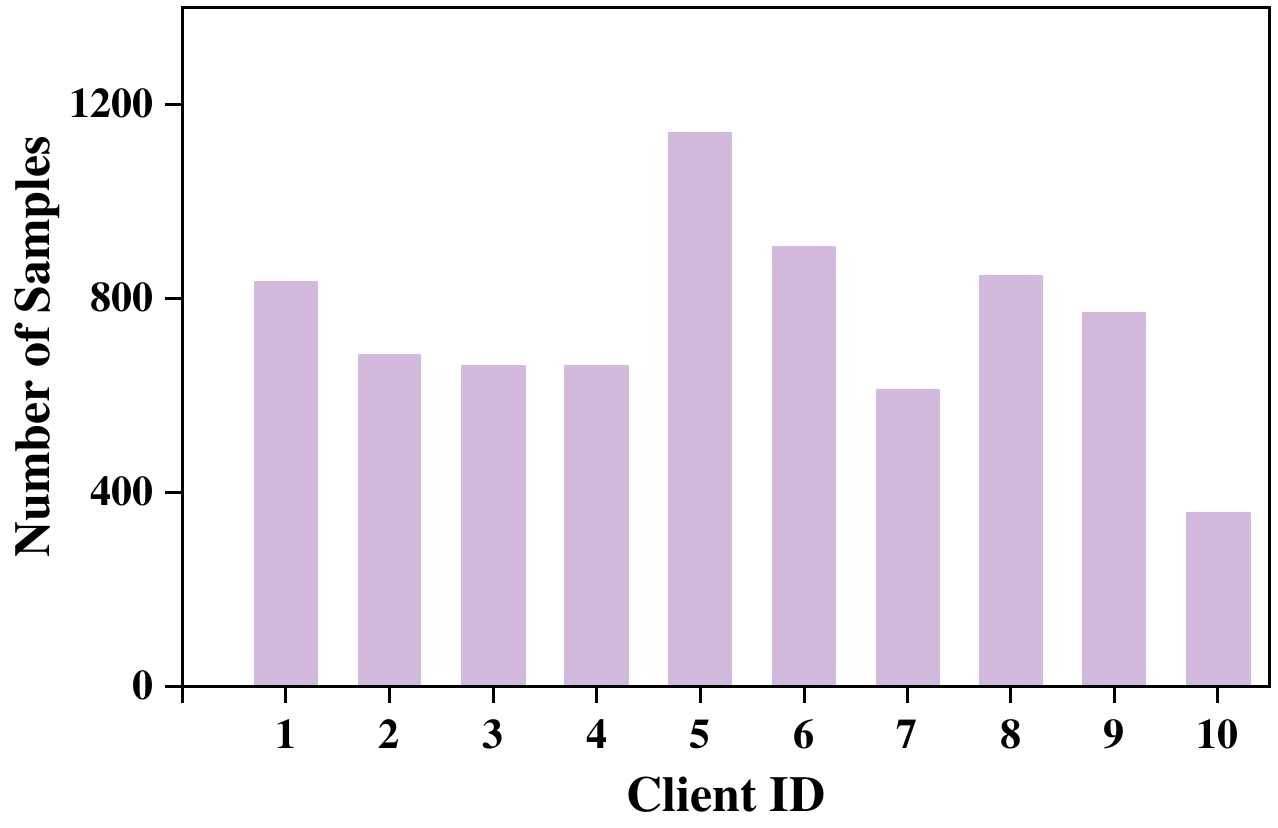}
        \label{fig:GSM8K_10.0}
    }\\ 
    \subfloat[SST-2 $\beta=0.5$]{
        \includegraphics[width=0.23\textwidth]{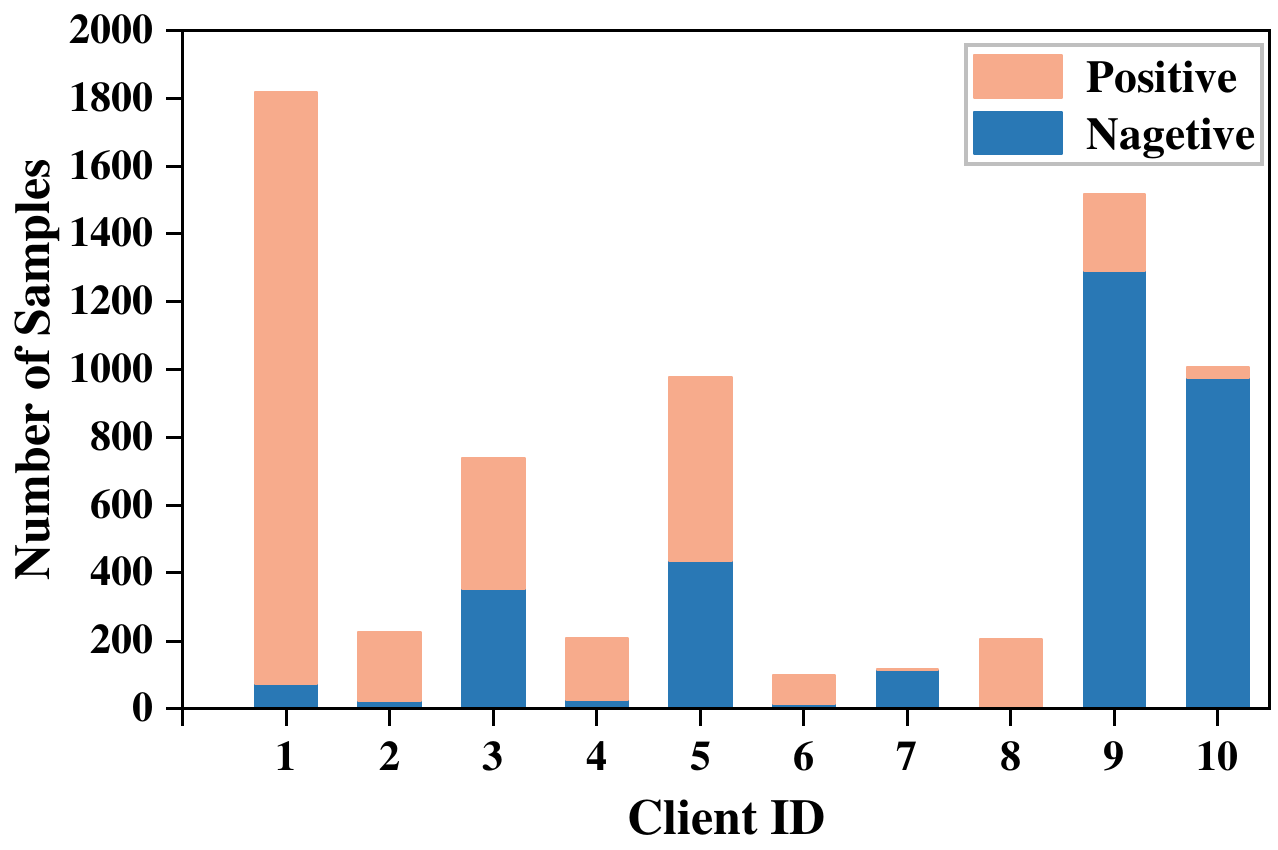}
        \label{fig:SST2_0.5}
    }\hfill
    \subfloat[SST-2 $\beta=1.0$]{
        \includegraphics[width=0.23\textwidth]{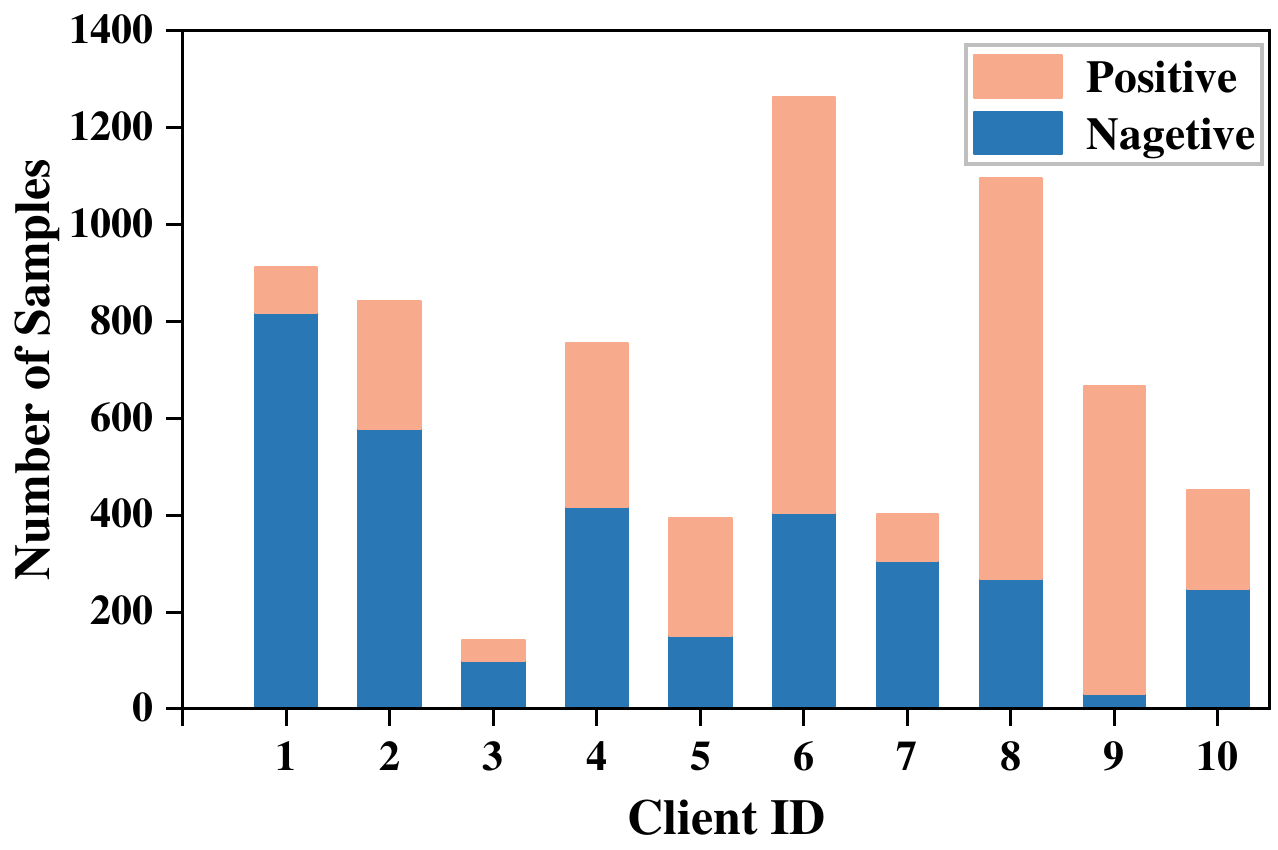}
        \label{fig:SST2_1.0}
    }\hfill
    \subfloat[SST-2 $\beta=5.0$]{
        \includegraphics[width=0.23\textwidth]{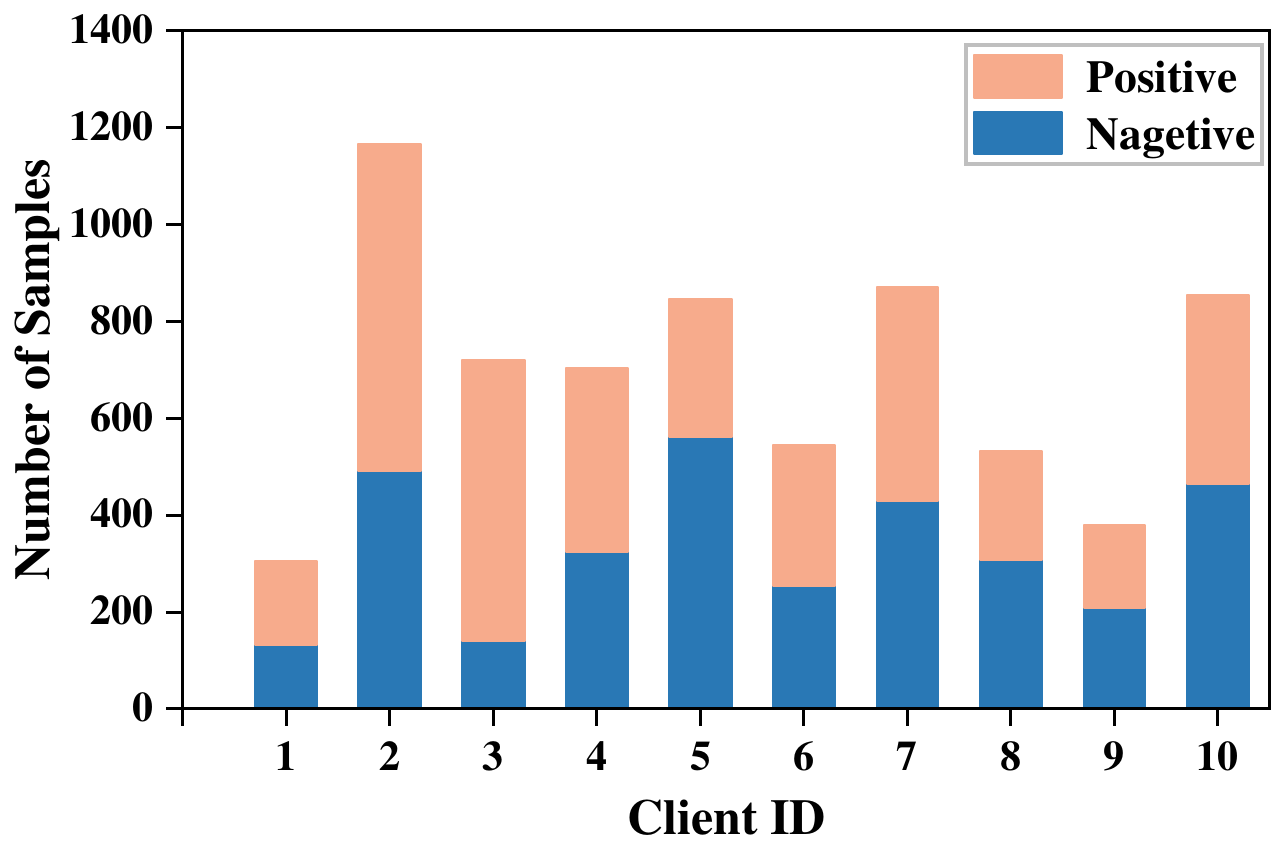}
        \label{fig:SST2_5.0}
    }\hfill
    \subfloat[SST-2 $\beta=10.0$]{
        \includegraphics[width=0.23\textwidth]{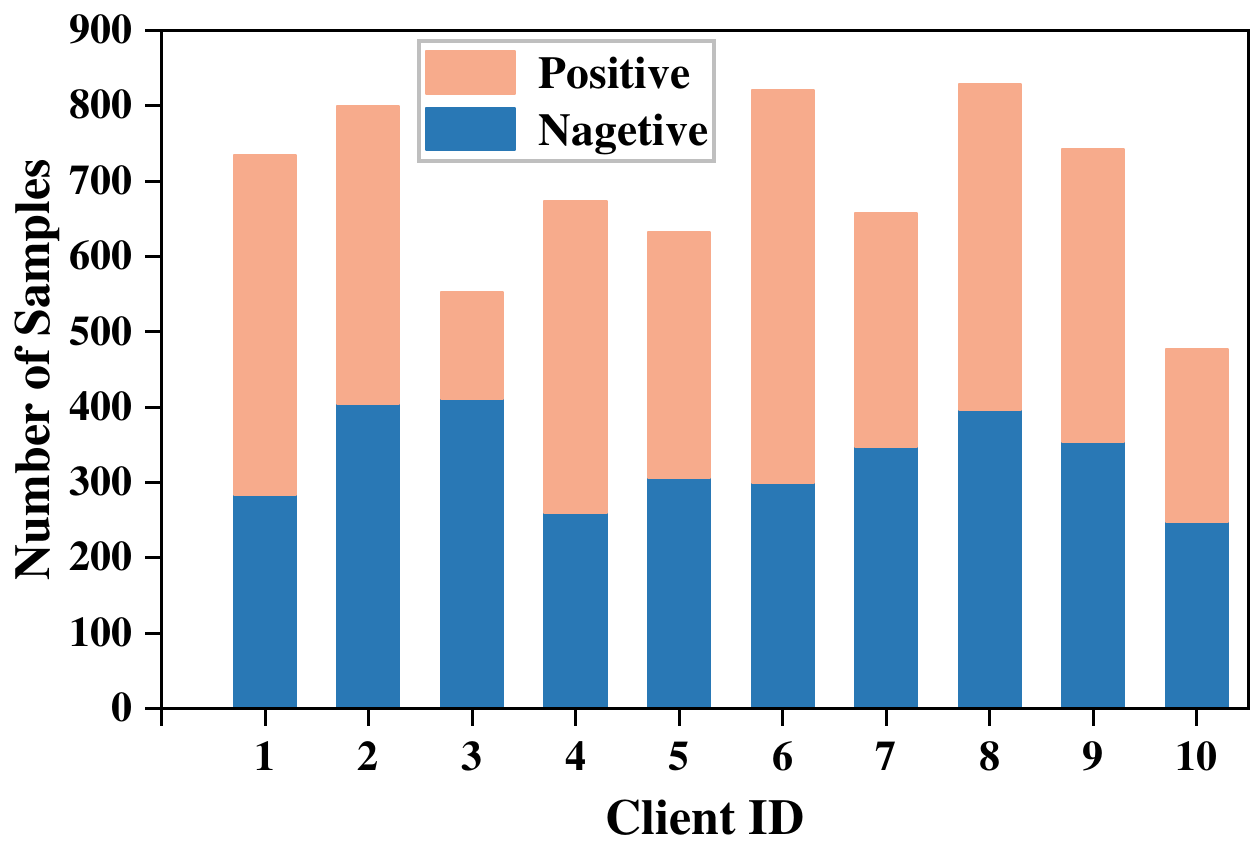}
        \label{fig:SST2_10.0}
    }\\ 
     \subfloat[BoolQ $\beta=0.5$]{
        \includegraphics[width=0.23\textwidth]{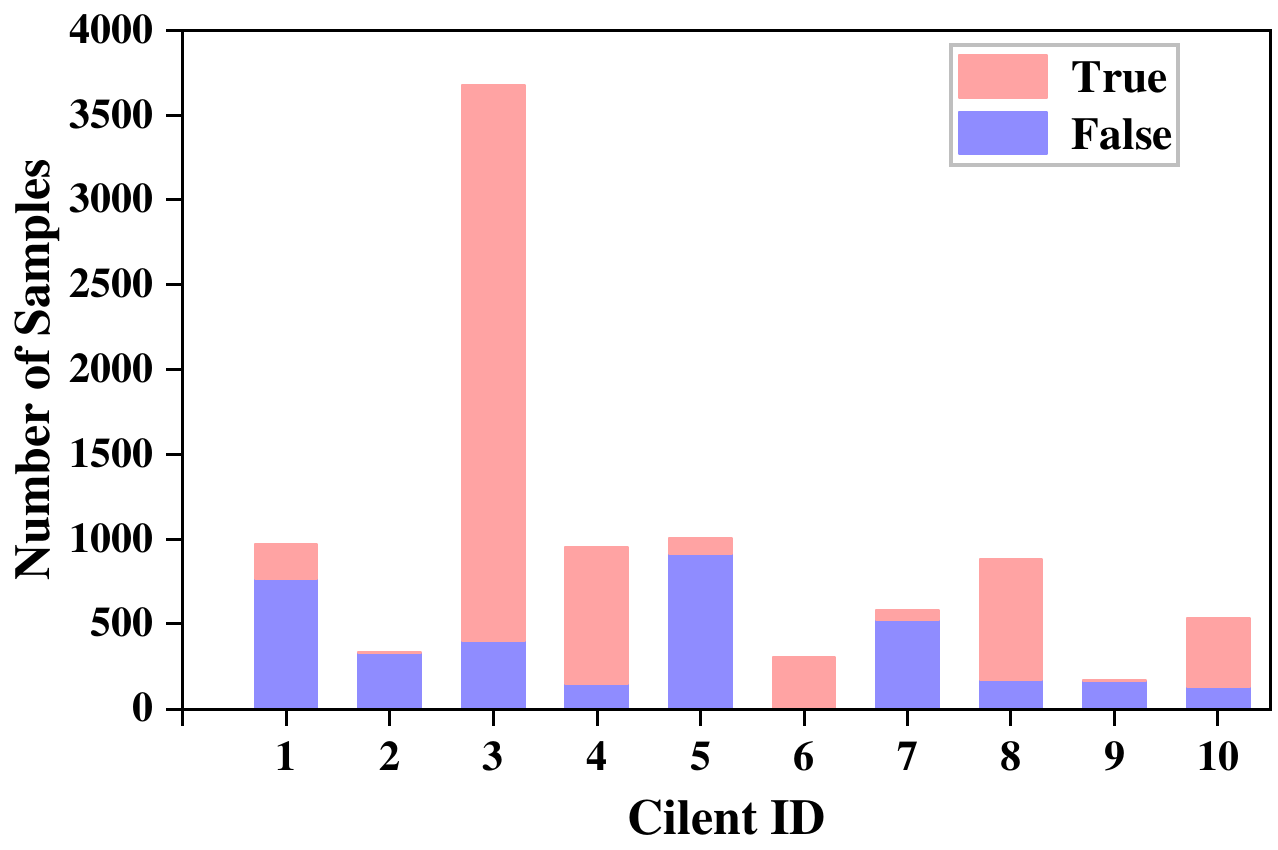}
        \label{fig:BoolQ_0.5}
    }\hfill
    \subfloat[BoolQ $\beta=1.0$]{
        \includegraphics[width=0.23\textwidth]{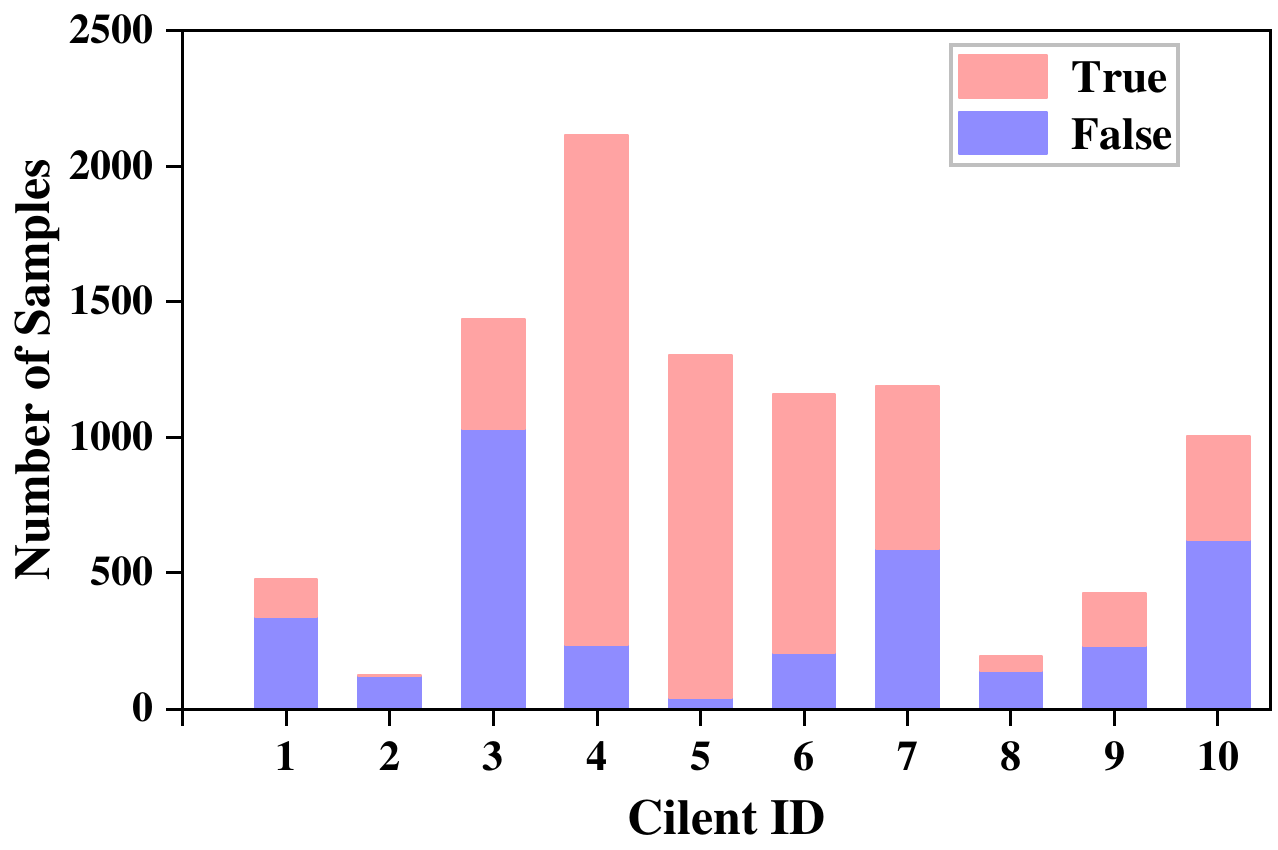}
        \label{fig:BoolQ_1.0}
    }\hfill
    \subfloat[BoolQ $\beta=5.0$]{
   \includegraphics[width=0.23\textwidth]{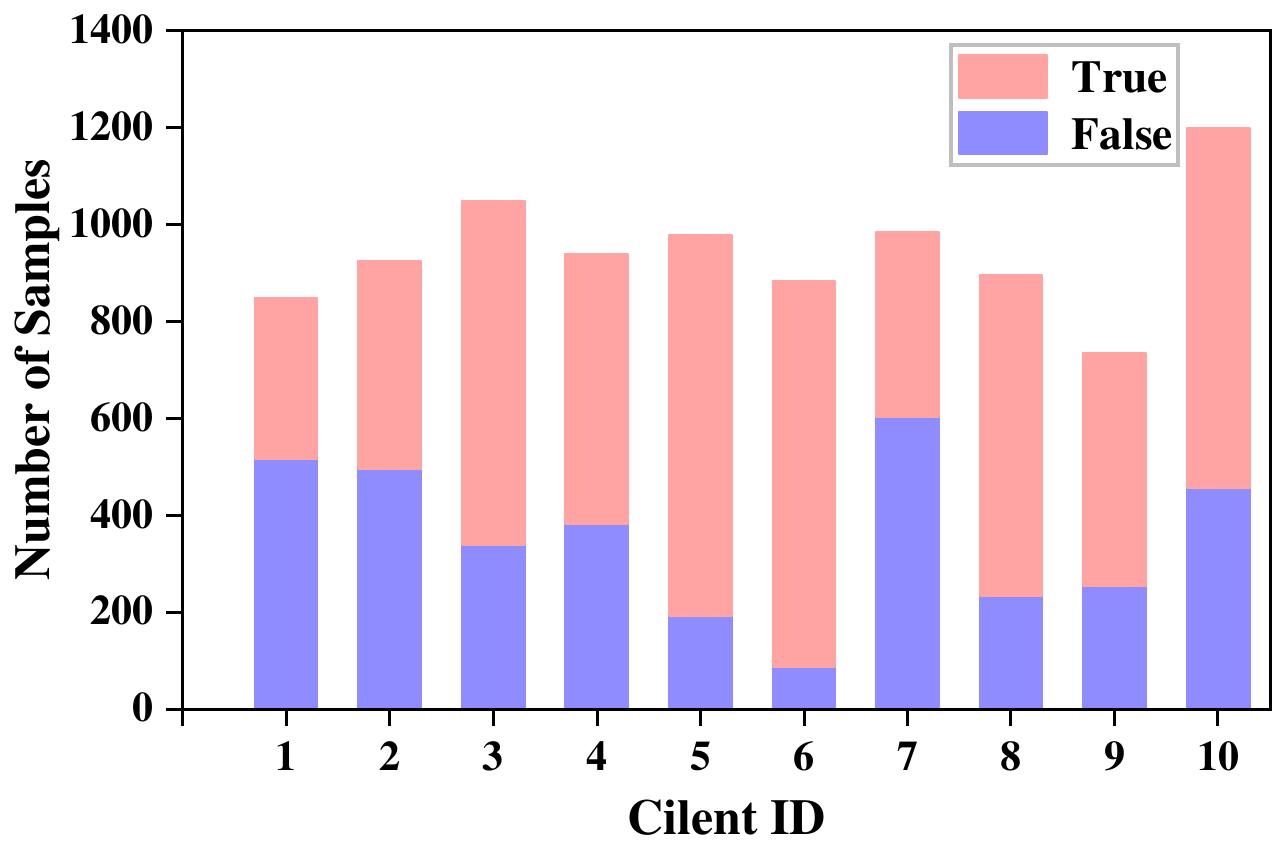}
        \label{fig:BoolQ_5.0}
    }\hfill
    \subfloat[BoolQ $\beta=10.0$]{
    \includegraphics[width=0.23\textwidth]{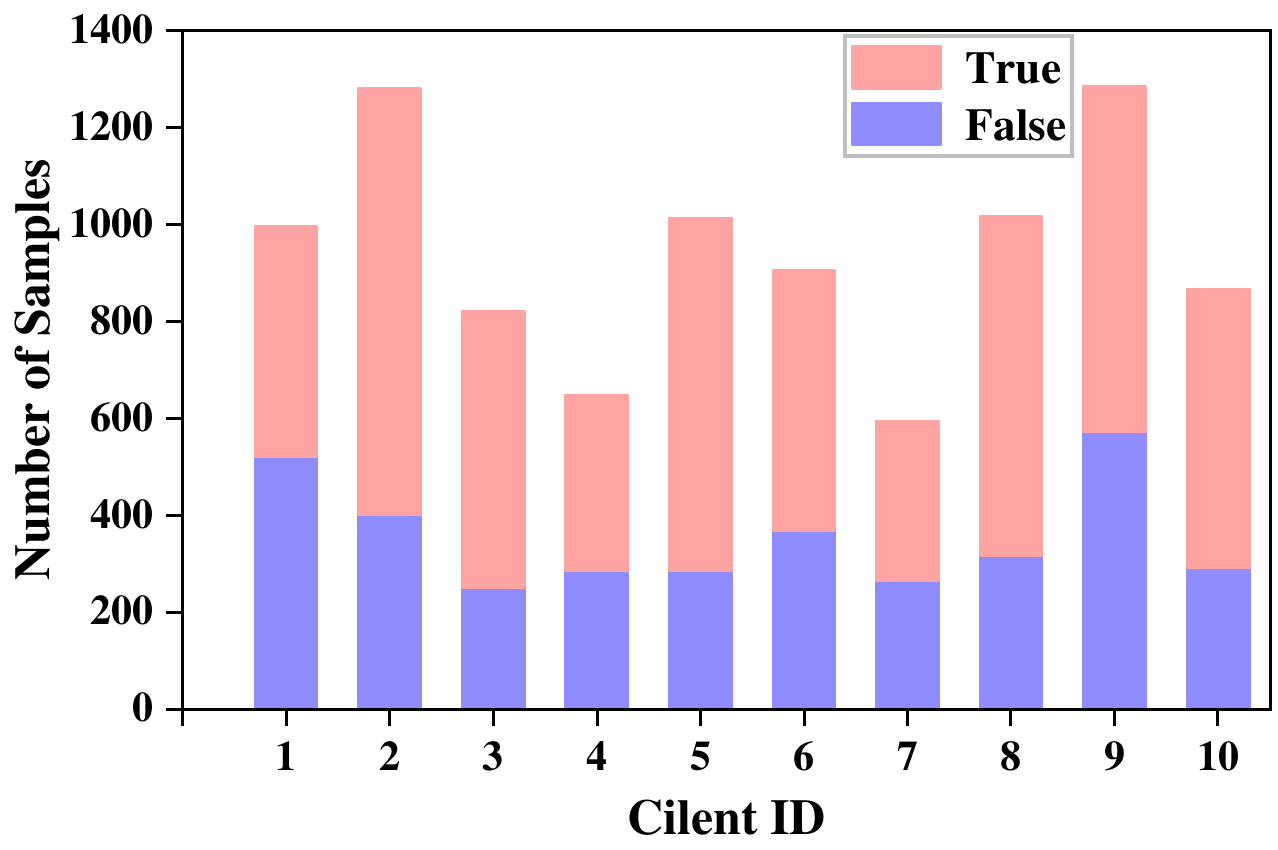}
        \label{fig:BoolQ_10.0}
    }
   
    \caption{Client Dirichlet Distribution Visualisations on GSM8K, SST-2 and BoolQ Datasets}
    \label{fig:Dirichlet}
\end{figure*}

We presented the Dirichlet distribution visualisations for 10 clients across the GSM8K, SST-2, and BoolQ datasets, as shown in the Figure \ref{fig:Dirichlet}.

\end{document}